\documentclass{article}

     \PassOptionsToPackage{numbers, compress}{natbib}

    \usepackage[final]{neurips_2023}

\usepackage[utf8]{inputenc} 
\usepackage[T1]{fontenc}    
\usepackage[colorlinks=true,linkcolor=black,citecolor=blue,urlcolor=blue,]{hyperref}       
\usepackage{url}            
\usepackage{booktabs}       
\usepackage{amsfonts}       
\usepackage{nicefrac}       
\usepackage{microtype}      
\usepackage{xcolor}         
\usepackage{graphicx}
\usepackage{amsmath}
\usepackage{bm}
\usepackage{amssymb}
\usepackage{wrapfig}
\usepackage{algorithm}
\usepackage{algorithmic}
\usepackage{subfigure}
\usepackage{multirow}

\title{Neural Multi-Objective Combinatorial Optimization with Diversity Enhancement}

\author{
Jinbiao Chen$^{1}$,
Zizhen Zhang$^{1}$,
Zhiguang Cao$^{2}$,
Yaoxin Wu$^{3}$,
Yining Ma$^{4}$,\\
\textbf{Te Ye}$^{1}$,
and \textbf{Jiahai Wang}$^{1,5,6,}$\thanks{Corresponding Author.}\\
$^1$School of Computer Science and Engineering, Sun Yat-sen University, P.R. China\\
$^2$School of Computing and Information Systems, Singapore Management University, Singapore\\
$^3$Department of Industrial Engineering \& Innovation Sciences,\\Eindhoven University of Technology, Netherlands\\
$^4$Department of Industrial Systems Engineering \& Management,\\National University of Singapore, Singapore\\
$^5$Key Laboratory of Machine Intelligence and Advanced Computing, Ministry of Education,\\Sun Yat-sen University, P.R. China\\
$^6$Guangdong Key Laboratory of Big Data Analysis and Processing, Guangzhou, P.R. China\\
\texttt{chenjb69@mail2.sysu.edu.cn},~\texttt{zhangzzh7@mail.sysu.edu.cn}\\
\texttt{zgcao@smu.edu.sg},~
\texttt{y.wu2@tue.nl},~
\texttt{yiningma@u.nus.edu}\\
\texttt{yete@mail2.sysu.edu.cn},~
\texttt{wangjiah@mail.sysu.edu.cn}
}

\begin{document}

\maketitle

\begin{abstract}

  Most of existing neural methods for multi-objective combinatorial optimization (MOCO) problems solely rely on decomposition, which often leads to repetitive solutions for the respective subproblems, thus a limited Pareto set. Beyond decomposition, we propose a novel neural heuristic with diversity enhancement (NHDE) to produce more Pareto solutions from two perspectives. On the one hand, to hinder duplicated solutions for different subproblems, we propose an indicator-enhanced deep reinforcement learning method to guide the model, and design a heterogeneous graph attention mechanism to capture the relations between the instance graph and the Pareto front graph. On the other hand, to excavate more solutions in the neighborhood of each subproblem, we present a multiple Pareto optima strategy to sample and preserve desirable solutions. Experimental results on classic MOCO problems show that our NHDE is able to generate a Pareto front with higher diversity, thereby achieving superior overall performance. Moreover, our NHDE is generic and can be applied to different neural methods for MOCO.

\end{abstract}

\section{Introduction}

Multi-objective combinatorial optimization (MOCO) has been extensively studied in the communities of computer science and operations research \cite{sai21,liu20}. It also commonly exists in many industries, such as transportation \cite{zaj21}, manufacturing \cite{tur20}, energy \cite{cui17}, and telecommunication \cite{fei17}. MOCO features multiple conflicting objectives based on NP-hard combinatorial optimization (CO), practical yet more complex. Rather than finding an optimal solution like in the single-objective optimization, MOCO pursues a set of Pareto-optimal solutions, called \emph{Pareto set}, to trade-off the multiple objectives. In general, a decent Pareto set is captured by both desirable convergence (optimality) and diversity.

Since exactly solving MOCO may require exponentially increasing computational time~\cite{ehr16}, the heuristic methods \cite{her21} have been favored in practice over the past few decades, which aim to yield an approximate Pareto set. Despite a relatively high efficiency, heuristic methods need domain-specific knowledge and massive iterative search. As \emph{neural CO} methods based on deep reinforcement learning (DRL) recently achieved notable success in CO problems such as routing, scheduling, and bin packing \cite{zha23,maz21,ben21,yan22}, a number of \emph{neural MOCO} methods have also been accordingly developed \cite{lik21,lin22,zha22,wuy22b}. Typically, parameterized as a deep model, Neural MOCO is able to automatically learn a heuristic (or policy), so as to directly construct near-optimal solutions in an end-to-end fashion, which takes much less computational time than the traditional ones.

Existing neural MOCO methods mostly decompose an MOCO problem into a series of single-objective CO subproblems and derive a Pareto set by solving them. However, while enjoying a favorable efficiency for neural MOCO, the sole decomposition is less effective in finding as many diverse solutions as possible, since constructing an optimal solution for the decomposed subproblems is always carried out independently, causing repetitive or duplicated ones for different subproblems.

\begin{figure}
    \centering
    \includegraphics[width=\textwidth]{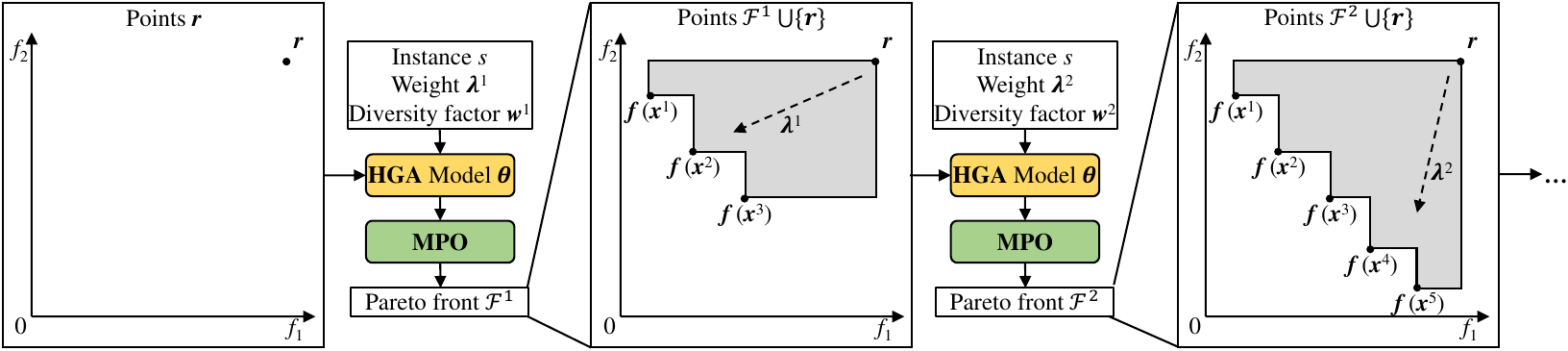}
    \caption{The framework of NHDE. For subproblem $i$, the heterogeneous graph attention (HGA) model takes instance $s$, \emph{points} $\mathcal{F}^{i-1} \cup \{\bm{r}\}$, weight $\bm{\lambda}^i$, and diversity factor $\bm{w}^i$ as inputs, and generates solutions to optimize the scalar objective and hypervolume (size of the gray area). More solutions are sampled and the Pareto front $\mathcal{F}^i$ is then efficiently updated based on multiple Pareto optima (MPO).}
    \label{fig:frame}
\end{figure}

To tackle this issue, we propose a neural heuristic with diversity enhancement (NHDE), as illustrated in Figure \ref{fig:frame}. Distinguished from existing neural methods, NHDE couples the decomposition with a comprehensive indicator to learn a policy that can produce diverse solutions across the subproblems while further improving the performance. Besides, for a given subproblem, multiple relevant solutions, rather than a single optimal solution with respect to the scalar objective, are found based on a proposed multiple Pareto optima (MPO) strategy, so as to further strengthen the diversity.

Our contributions are summarized as follows. (1) We propose an indicator-enhanced DRL method. To encourage the deep model to generate different yet diverse solutions for decomposed subproblems, NHDE inputs the Pareto front composed of the preceding solutions and introduces an indicator comprehensively measuring convergence and diversity in the reward. (2) We design a heterogeneous graph attention model to effectively capture the correlation between an instance graph and its Pareto front graph. (3) We present a multiple Pareto optima (MPO) strategy to further identify more relevant solutions in the neighborhood of each subproblem and efficiently update the Pareto front. (4) We deploy our NHDE with two different types of neural MOCO methods to demonstrate its versatility. Experimental results based on various MOCO problems show that our NHDE outperforms state-of-the-art neural baselines, especially with significant improvement in the diversity.

\section{Related works}

\textbf{Exact and heuristic methods for MOCO.} Exact methods \cite{ehr16,ber22} for MOCO can attain the accurate Pareto set, but their computational time may grow exponentially, rendering them less practical. As an alternative, heuristic methods such as multi-objective evolutionary algorithms (MOEAs) have gained widespread attention in practice. Dominance-based NSGA-II \cite{deb02}, decomposition-based MOEA/D \cite{zha07}, and indicator-based SMS-EMOA \cite{beu07} are three typical paradigms of MOEAs. In these MOEAs, local search, a crucial technique specialized to the target CO, is usually employed \cite{jas02,shi20,shi22}.

\textbf{Neural CO.} In the past few years, neural \emph{construction} methods \cite{vin15,bel17,naz18} were proposed to rapidly yield high-quality solutions in an end-to-end fashion. A well-known representative is \emph{Attention Model} (AM) \cite{koo19}, which was developed based on the \emph{Transformer} architecture \cite{vas17}. Then, AM inspired a large number of subsequent works \cite{xin21,kwo21,kim22,bij22,che22,zha23b,zhou2023towards}, which further boosted the performance. Among them, the policy optimization with multiple optima (POMO)~\cite{kwo20}, leveraging the solution and problem symmetries, is recognized as a prominent approach. Besides, the other line of works,  known as neural \emph{improvement} methods \cite{che19,luh20,wuy22,may21,may22}, exploited DRL to assist the iterative improvement process from an initial but complete solution, following a learn-to-improve paradigm.

\textbf{Neural MOCO.} Decomposition is a mainstream scheme in learning-based methods for multi-objective optimization \cite{lin22b,nav21,lin19}. An MOCO problem can be decomposed into a series of single-objective CO problems and then solved by neural construction methods to approximate the Pareto set. A couple of preliminary works trained multiple deep models with transfer learning \cite{lik21,wuh20}, where each deep model coped with one subproblem. Evolutionary learning \cite{sha23,zha21} was introduced to evolve deep models to further improve the performance. Instead of training multiple deep models for preset weights, preference-conditioned multi-objective combinatorial optimization (PMOCO) \cite{lin22} and Meta-DRL (MDRL) \cite{zha22}, both of which only trained one deep model, were more flexible and practical. For a given weight vector, the former used a hypernetwork to derive the decoder parameters to solve the corresponding subproblem, while the latter rapidly fine-tuned a pre-trained meta-model to solve the corresponding subproblem. However, those solely decomposition-based neural MOCO methods are limited in the diversity with respect to the solutions of the Pareto set, since some subproblems may lead to duplicated solutions, especially when they are solved independently.

\section{Preliminary}

\subsection{MOCO}

An MOCO problem with $M$ objectives can be expressed as $\underset{\bm{x} \in \mathcal{X} }{\min}\; \bm{f}(\bm{x}) = (f_1(\bm{x}), f_2(\bm{x}), \dots, f_M(\bm{x}))$, 
where $\mathcal{X}$ is a set of discrete decision variables.

\textbf{Definition 1 (Pareto dominance).} A solution $\bm{x}^1 \in \mathcal{X}$ dominates another solution $\bm{x}^2 \in \mathcal{X}$ ($\bm{x}^1 \prec \bm{x}^2$), if and only if $f_i(\bm{x}^1) \leq f_i(\bm{x}^2), \forall i \in \{1, \dots, M\}$ and $\exists j \in \{1, \dots, M\}, f_j(\bm{x}^1) < f_j(\bm{x}^2) $.

\textbf{Definition 2 (Pareto optimality).} A solution $\bm{x}^* \in \mathcal{X}$ is Pareto-optimal if it is not dominated by any other solution, i.e., $\nexists \bm{x}' \in \mathcal{X}$ such that $\bm{x}' \prec \bm{x}^*$.

\textbf{Definition 3 (Pareto set/front).} MOCO aims to uncover a Pareto set, comprising all Pareto optimal solutions $\mathcal{P} = \{\bm{x}^* \in \mathcal{X}~|~ \nexists \bm{x}' \in \mathcal{X}: \bm{x}' \prec \bm{x}^* \}$. The Pareto front  $\mathcal{F} = \{\bm{f}(\bm{x}) ~|~ \bm{x} \in \mathcal{P} \}$ corresponds to the objective values of Pareto set, with each $\bm{f}(\bm{x})$ referred to as a \emph{point} in the objective space.

\subsection{Decomposition}

For MOCO, decomposition \cite{zha07} is a prevailing scheme due to its flexibility and effectiveness. An MOCO problem can be decomposed into $N$ subproblems with $N$ weights. Each subproblem is a single-objective CO problem via scalarization $g(\bm{x}|\bm{\lambda})$  with a weight $\bm{\lambda}\in \mathcal{R}^M$ satisfying $\lambda_m\geq0$ and $\sum_{m=1}^{M}\lambda_m=1$. The Pareto set then can be derived by solving the $N$ subproblems.

The simplest yet effective decomposition approach is the weighted sum (WS). It uses the linear scalarization of $M$ objectives, which hardly raises the complexity of the subproblems, as follows,
\begin{equation}
	\underset{\bm{x} \in \mathcal{X}}{\min}\; g_{\rm{ws}}(\bm{x}|\bm{\lambda}) = \sum_{m=1}^{M}\lambda_m f_m(\bm{x}).
\end{equation}

\subsection{Indicator}

Hypervolume (HV) is a mainstream indicator to measure performance, as it can comprehensively assess the convergence and diversity without the exact Pareto front \cite{aud21}. For a Pareto front $\mathcal{F}$ in the objective space, ${\rm{HV}}_{\bm{r}}(\mathcal{F})$ with respect to a fixed reference point $\bm{r} \in \mathcal{R}^M$ is defined as follows,
\begin{equation}
	{\rm{HV}}_{\bm{r}}(\mathcal{F})=\mu \left( \underset{\bm{f}(\bm{x}) \in \mathcal{F}}{\bigcup}[\bm{f}(\bm{x}), \bm{r}] \right),
\end{equation}
where $\mu$ is the Lebesgue measure, i.e., $M$-dimensional volume, and $[\bm{f}(\bm{x}), \bm{r}]$ is a $M$-dimensional cube, i.e., $[\bm{f}(\bm{x}), \bm{r}] = [f_1(\bm{x}), r_1] \times \dots \times [f_M(\bm{x}), r_M]$.

A 2-dimensional example with 5 \emph{points} in the objective space is depicted in Figure \ref{fig:frame}, where $\mathcal{F}=\{\bm{f}(\bm{x}^1),\bm{f}(\bm{x}^2),\bm{f}(\bm{x}^3),\bm{f}(\bm{x}^4),\bm{f}(\bm{x}^5)\}$. ${\rm{HV}}_{\bm{r}}(\mathcal{F})$ is equal to the size of the gray area, and finally normalized into $[0,1]$. All methods share the same reference point $\bm{r}$ for a problem (see Appendix A).

\section{Methodology}

Our \emph{neural heuristic with diversity enhancement} (NHDE) exploits indicator-enhanced DRL to produce diverse solutions across different subproblems and leverages a multiple Pareto optima (MPO) strategy to find multiple neighbor solutions for each subproblem, as illustrated in Figure \ref{fig:frame}. Specifically, an MOCO problem is decomposed into $N$ single-objective subproblems with $N$ weights, which are solved dependently by a unified heterogeneous graph attention (HGA) model $\bm{\theta}$. For each subproblem $i$, its features together with the current Pareto front $\mathcal{F}$ (\emph{points} in the objective space) constituted by preceding solutions are input to model $\bm{\theta}$, which is guided by the scalar objective with the HV indicator. Then, MPO is utilized to sample multiple solutions and efficiently update $\mathcal{F}$.

\subsection{Indicator-enhanced DRL}

Given a problem instance $s$, we sequentially solve its subproblem $i \in \{1, \dots, N\}$, each associated with weight $\bm{\lambda}^i$. Let $\bm{\pi}^i\!=\!\{\pi_1^i,\dots,\pi_{T}^i\}$ denote the obtained solution\footnote{In this sub-section, we consider the construction of only one solution in each step for better readability; however, we note that multiple solutions can be sampled and our formulation would work in a similar manner.}
at step $i$, and let $\mathcal{F}^{i}$ be the Pareto front yielded by solutions from subproblem 1 to $i$.
In each step $i$, we select up to $K$ top \emph{points} $\bm{f}(\bm{\pi}) \in \mathcal{F}^{i-1}$ from the Pareto front at step $i\!-\!1$ with the ranking determined by the scalar objective $g(\bm{\pi}|s,\bm{\lambda}^i)$ with respect to the new given weight $\bm{\lambda}^i$. The corresponding scalar objective and the induced surrogate landscape $\tilde{\mathcal{F}}^{i-1} \subseteq \mathcal{F}^{i-1}$ based on those selected solutions are treated as the policy network inputs (see Figure \ref{fig:model}), so as to construct a new solution $\bm{\pi}^i$ and yield a new $\mathcal{F}^{i}$.

The construction of the solution $\bm{\pi}^i$ with length $T$ for each subproblem $i$ can be cast as a Markov decision process. In particular, 1) the \emph{state} includes the weight $\bm{\lambda}^i$, user-defined diversity factor $\bm{w}^i\in \mathcal{R}^2$ satisfying $w_1^i, w_2^i\geq0$ and $w_1^i+w_2^i=1$, partial solution $\bm{\pi}_{1:t-1}^i$, instance $s$, and $\tilde{\mathcal{F}}_{\bm{r}}^{i-1}$, where $\tilde{\mathcal{F}}_{\bm{r}}^{i-1}=\tilde{\mathcal{F}}^{i-1}\cup\{\bm{r}\}$ incorporates the aforementioned surrogate landscape at step $i$ and the given reference \emph{point} $\bm{r}$; 2) the \emph{action} is to add a node $\pi_t^i$ into $\bm{\pi}_{1:t-1}^i$; 3) the \emph{state transition} transforms $\bm{\pi}_{1:t-1}^i$ to $\bm{\pi}_{1:t}^i$, denoted as $\bm{\pi}_{1:t}^i=\{\bm{\pi}_{1:t-1}^i,\bm{\pi}_t^i\}$; 
4) the \emph{reward} is defined as $R^i=-w_1^i \times g(\bm{\pi}^i|s,\bm{\lambda}^i)+ w_2^i \times {\rm{HV}}_{\bm{r}}(\tilde{\mathcal{F}}^{i-1}\cup\{\bm{f}(\bm{\pi}^i)\})$, where we introduce the hypervolume ${\rm{HV}}_{\bm{r}}(\tilde{\mathcal{F}}^{i-1}\cup\{\bm{f}(\bm{\pi}^i)\})$ to guide the search; and 5) the stochastic \emph{policy} generating the solution $\bm{\pi}^i$ is expressed as $P(\bm{\pi}^i|s,\tilde{\mathcal{F}}_{\bm{r}}^{i-1},\bm{\lambda}^i,\bm{w}^i)=\prod_{t=1}^{T} P_{\bm{\theta}}(\pi_t^i|\bm{\pi}_{1:t-1}^i,s,\tilde{\mathcal{F}}_{\bm{r}}^{i-1},\bm{\lambda}^i,\bm{w}^i)$, with the probability of node selection $P_{\bm{\theta}}(\pi_t^i|\bm{\pi}_{1:t-1}^i,s,\tilde{\mathcal{F}}_{\bm{r}}^{i-1},\bm{\lambda}^i,\bm{w}^i)$ parameterized by a deep model $\bm{\theta}$.

We would like to note that our NHDE is generic, and can directly integrate the base model $\bm{\theta}$ with the existing decomposition-based neural MOCO methods. We demonstrate this property by applying it to two state-of-the-art methods, PMOCO \cite{lin22} and MDRL \cite{zha22}, denoted as NHDE-P and NHDE-M, respectively. Given $\bm{\lambda}^i$ and $\bm{w}^i$ as inputs, NHDE-P uses a hypernetwork to generate the decoder parameters of the model $\bm{\theta}(\bm{\lambda}^i,\bm{w}^i)$, while NHDE-M fine-tunes the pre-trained meta model $\bm{\theta}_{\rm{meta}}$ with a few steps to address the corresponding subproblem. More details are presented in Appendix B.

\subsection{Heterogeneous graph attention}

\begin{figure}
    \centering
    \includegraphics[width=\textwidth]{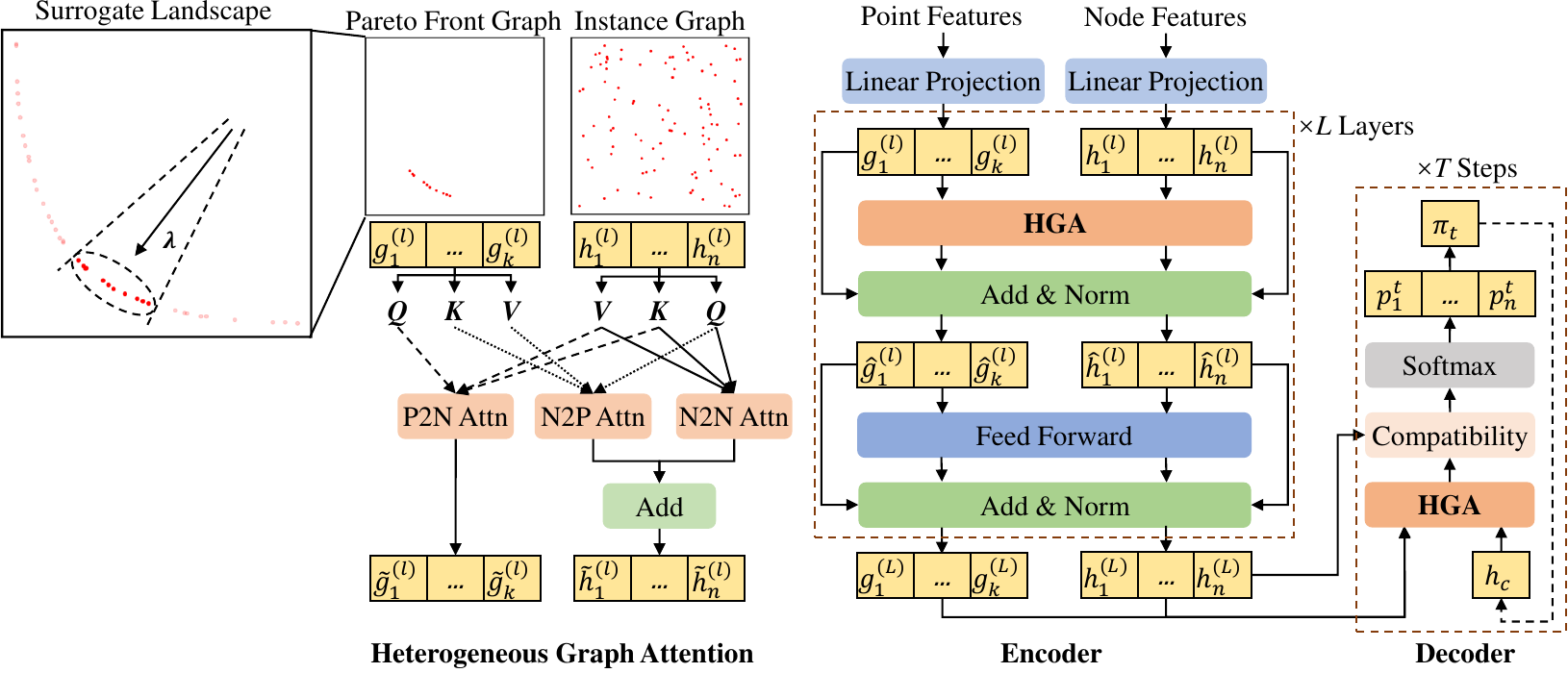}
    \caption{Illustration of the proposed heterogeneous graph attention (HGA) model.}
    \label{fig:model}
\end{figure}

To effectively solve subproblem $i$, the model
should jointly capture the representations of 
both the instance's node graph and the Pareto front's \emph{point} graph. 
Based on the encoder-decoder structure, we thus design a heterogeneous graph attention (HGA) model to correlate the two heterogeneous graphs as depicted in Figure \ref{fig:model}. In the following section, we omit superscript $i$ for better readability.

\textbf{Encoder.} Given an instance graph $s$ containing $n$ nodes with $Z$-dimensional features and a Pareto front graph $\tilde{\mathcal{F}}_{\bm{r}}^{i-1}$ containing $k$ \emph{points} ($k \leq K+1$) with $M$-dimensional features, their initial embeddings $\bm{h}_1^{(0)}, \dots, \bm{h}_n^{(0)} \in R^d$ and $\bm{g}_1^{(0)}, \dots, \bm{g}_k^{(0)} \in R^d$ are yielded by a linear projection with trainable parameters $W_h$ and $W_g$, respectively, and $d$ is empirically set to 128. The eventual embeddings $\bm{h}^{(L)}_1, \dots, \bm{h}^{(L)}_n$ and $\bm{g}^{(L)}_1, \dots, \bm{g}^{(L)}_k$ are derived by further passing through $L=6$ attention layers. Each attention layer is composed of a multi-head HGA layer with $Y=8$ heads and a fully connected feed-forward sublayer. For layer $l \in \{1,\dots, L\}$, HGA computes the representations $\tilde{\bm{h}}_u^{(l)}$ and $\tilde{\bm{g}}_u^{(l)}$ across the heterogeneous graphs, which are used to update the embeddings $\bm{h}_u^{(l)}$ and $\bm{g}_u^{(l)}$. Skip-connection \cite{hek16} and batch normalization \cite{iof15} are both adopted in each sublayer, as follows,
\begin{equation}
	\bm{h}_u^{(l)}={\rm{BN}}(\hat{\bm{h}}_u+{\rm{FF}}(\hat{\bm{h}}_u)),~\hat{\bm{h}}_u={\rm{BN}}(\bm{h}_u^{(l-1)}+\tilde{\bm{h}}_u^{(l)}),\quad \forall u \in \{1,\dots, n\},
\end{equation}
\begin{equation}
	\bm{g}_u^{(l)}={\rm{BN}}(\hat{\bm{g}}_u+{\rm{FF}}(\hat{\bm{g}}_u)),~\hat{\bm{g}}_u={\rm{BN}}(\bm{g}_u^{(l-1)}+\tilde{\bm{g}}_u^{(l)}),\quad \forall u \in \{1,\dots, k\}.
\end{equation}

\textbf{Decoder.} The decoder, composed of a multi-head HGA layer and a \emph{compatibility} layer, autoregressively constructs a solution according to the probability distribution with $T$ steps. At decoding step $t \in \{1,..., T\}$, the \emph{glimpse} $\bm{q}_c$ of the \emph{context} embedding $\bm{h}_c$ (see Appendix C) is computed by the HGA layer. Then, the \emph{compatibility} $\bm{\alpha}$ is calculated as follows,
\begin{equation}
	\alpha_{u}=\left\{
	\begin{array}{lcl}
		-\infty, & & {\rm{node}}~u~{\rm{is~masked}}\\
		C \cdot \tanh (\frac{\bm{q}_c^T (W^K \bm{h}_{u}^{(L)})}{\sqrt{d/Y}}), & & \rm{otherwise}
	\end{array} \right.
\end{equation}    
where $C$ is set to 10 \cite{koo19}. Finally, softmax is employed to calculate the selection probability distribution $P_{\bm{\theta}}(\bm{\pi}|s,\tilde{\mathcal{F}}_{\bm{r}}^{i-1},\bm{\lambda}^i,\bm{w}^i)$ for nodes, i.e., $P_{\bm{\theta}}(\pi_t|\bm{\pi}_{1:t-1},s,\tilde{\mathcal{F}}_{\bm{r}}^{i-1},\bm{\lambda}^i,\bm{w}^i)=\text{Softmax}(\bm{\alpha})$.

\textbf{HGA.}
The HGA layer in the encoder captures three key relations between the node graph and the \emph{point} graph. The first, \emph{node-to-node} $\alpha_{uv}^{hh}$, indicates each node's attention towards others within the same instance to construct a promising solution. The second, \emph{node-to-point} $\alpha_{uv}^{hg}$, suggests each node's attention to \emph{points}, guiding the constructed solutions distinct from the existing ones in the current Pareto front. The third, \emph{point-to-node} $\alpha_{uv}^{gh}$, indicates each \emph{point}'s attention to nodes, facilitating the learning of the mapping from a solution to its objective values. We disregard the less meaningful \emph{point-to-point} attention. Concretely, Eq. (\ref{equ8}) defines the above three attention scores, which are separately normalized as $\tilde{\alpha}_{uv}^{hh}$, $\tilde{\alpha}_{uv}^{hg}$, and $\tilde{\alpha}_{uv}^{hg}$ by softmax. Then, $\tilde{\bm{h}}_u$ and $\tilde{\bm{g}}_u$ are computed by Eq. (\ref{equ9}).
\begin{equation}
\label{equ8}
	\alpha_{uv}^{hh}=\frac{(W^Q_h \bm{h}_u)^T (W^K_h \bm{h}_v)}{\sqrt{d/Y}},~ \alpha_{uv}^{hg}=\frac{(W^Q_h \bm{h}_u)^T (W^K_g \bm{g}_v)}{\sqrt{d/Y}},~ \alpha_{uv}^{gh}=\frac{(W^Q_g \bm{g}_u)^T (W^K_h \bm{h}_v)}{\sqrt{d/Y}}.
\end{equation}
\begin{equation}
\label{equ9}
\tilde{\bm{h}}_u=\sum_{v=1}^{n}\tilde{\alpha}_{uv}^{hh}W^V_h \bm{h}_v+\sum_{v=1}^{k}\tilde{\alpha}_{uv}^{hg}W^V_g \bm{g}_v,~ \tilde{\bm{g}}_u=\sum_{v=1}^{n}\tilde{\alpha}_{uv}^{gh}W^V_h \bm{h}_v.
\end{equation}
Finally, as for the multi-head attention, $\tilde{\bm{h}}_u$ and $\tilde{\bm{g}}_u$ are further computed as follows,
\begin{equation}
\label{equ10}
	\tilde{\bm{h}}_u=W^O_h {\rm{Concat}}(\tilde{\bm{h}}_{u,1},\dots,\tilde{\bm{h}}_{u,Y}),~ \tilde{\bm{g}}_u=W^O_g {\rm{Concat}}(\tilde{\bm{g}}_{u,1},\dots,\tilde{\bm{g}}_{u,Y}),
\end{equation}
where $\tilde{\bm{h}}_{u,y}$ and $\tilde{\bm{g}}_{u,y}$ for head $y \in \{1,..., Y\}$ are obtained according to Eq. (\ref{equ9}). In the multi-head HGA, $W^Q_h$, $W^K_h$, $W^V_h$, $W^O_h$, $W^Q_g$, $W^K_g$, $W^V_g$, and $W^O_g$ are independent trainable parameters. Similarly, in the decoder, the glimpse $\bm{q}_c$ is calculated by the context embedding $\bm{h}_c$ with the addition of the \emph{context-to-node} and \emph{context-to-point} attention, i.e., replacing $\bm{h}_u$ with $\bm{h}_c$ in the Eq. (\ref{equ8}--\ref{equ10}).

\subsection{Multiple Pareto optima strategy}
Contrary to single-objective problems that focus on a single optimal solution, MOCO problems involve a series of Pareto-optimal solutions. In light of this, we introduce a multiple Pareto optima (MPO) strategy to uncover multiple solutions for each subproblem by leveraging the Pareto optimality.

When solving subproblem $i$, more than one solution can be attained by sampling, e.g., sampling with multiple start nodes as did in POMO \cite{kwo20}. In this case, $\mathcal{F}^i={\rm{MPO}}(\mathcal{F}^{i-1} \cup \mathcal{G}^i)$, where $\mathcal{G}^i$ contains all the candidate \emph{points} (find by sampling) to be introduced in the new Pareto front and ${\rm{MPO}}(\cdot)$ is an operator that updates the Pareto front. However, as the complexity of ${\rm{MPO}}(\mathcal{F}^{i-1} \cup \mathcal{G}^i)$ is $O((|\mathcal{F}^{i-1}|+|\mathcal{G}^i|)|\mathcal{G}^i|)$, it may take a relatively long time, especially when there are thousands of \emph{points} in $\mathcal{F}^{i-1}$ and $\mathcal{G}^i$. Thus, we suggest an efficient update mechanism executed on the surrogate Pareto fronts, $\mathcal{F}^i={\rm{MPO}}(\tilde{\mathcal{F}}^{i-1} \cup \tilde{\mathcal{G}}^i)$, where $\tilde{\mathcal{G}}^i \subset \mathcal{G}^i$ includes at most $J$ (usually setting $J > K$) best \emph{points} selected from $\bm{f}(\bm{\pi}) \in \mathcal{G}^i$ according to $g(\bm{\pi}|s,\bm{\lambda}^i)$. The complexity of ${\rm{MPO}}(\tilde{\mathcal{F}}^{i-1} \cup \tilde{\mathcal{G}}^i)$ is then reduced to $O((K+J)J)$, which is able to curtail the overall solving time in practice.

\subsection{Training and inference}

Our NHDE can be applied to different decomposition-based DRL methods, e.g., PMOCO \cite{lin22} and MDRL \cite{zha22}, and the training algorithm is easy to adapt with slight adjustments, where the one for NHDE-P is presented in Algorithm \ref{train-NHDE-P}. The key points include three aspects. (1) Multiple weights are sampled to train the same instance (Line 6), since the solving processes for those subproblems are dependent. (2) The HV indicator is adopted in the reward (Line 10). (3) Multiple Pareto-optimal solutions are preserved via MPO (Line 15). Note that when training with a batch, the \emph{point} sets with different sizes of the instances are padded with repetitive reference points, which are masked in the attention, based on the maximum size. The training algorithm of NHDE-M is given in Appendix E.

\begin{algorithm}[!t]
	\caption{Training algorithm of NHDE-P}
	\label{train-NHDE-P}
 	\begin{algorithmic}[1]
 	    \STATE \textbf{Input:} weight distribution $\Lambda$, diversity-factor distribution $\mathcal{W}$, instance distribution $\mathcal{S}$, number of training steps $E$, number of sampled weights per step $N'$, batch size $B$, instance size $n$
    	\STATE Initialize the model parameters $\bm{\theta}$
		\FOR{$e = 1$ to $E$}
			\STATE $s_i \sim \textbf{SampleInstance}(\mathcal{S}) \quad \forall i \in \{1,\cdots,B\}$
			\STATE Initialize $\mathcal{F}_i \leftarrow \emptyset \quad \forall i$
			\FOR{$n' = 1$ to $N'$}
			   \STATE $\bm{\lambda} \sim \textbf{SampleWeight}(\Lambda)$
			   \STATE $\bm{w} \sim \textbf{SampleDiversityFactor}(\mathcal{W})$
			   \STATE $\bm{\pi}_{i}^{j} \sim \textbf{SampleSolution}(P_{\bm{\theta}(\bm{\lambda},\bm{w})}(\cdot|s_i,\tilde{\mathcal{F}}_{\bm{r},i},\bm{\lambda},\bm{w})) \quad \forall i \in \{1,\cdots,B\} \quad \forall j \in \{1,\cdots,n\}$
			   \STATE $R_i^j \leftarrow -w_1 g(\bm{\pi}_i^j|s_i,\bm{\lambda})+ w_2 {\rm{HV}}_{\bm{r}}(\tilde{\mathcal{F}}_i\cup\{\bm{f}(\bm{\pi}_i^j)\}) \quad \forall i,j$
			   \STATE $b_i \leftarrow \frac{1}{n} \sum_{j=1}^{n} (-R_i^j) \quad \forall i$
			   \STATE $\nabla \mathcal{J}(\bm{\theta}) \leftarrow \frac{1}{Bn} \sum_{i=1}^{B} \sum_{j=1}^{n} [ (-R_i^j - b_i) \nabla_{\bm{\theta}(\bm{\lambda},\bm{w})} \log P_{\bm{\theta}(\bm{\lambda},\bm{w})}(\bm{\pi}_{i}^{j}|s_i,\tilde{\mathcal{F}}_{\bm{r},i},\bm{\lambda},\bm{w})]$
			   \STATE $\bm{\theta} \leftarrow \textbf{Adam}(\bm{\theta},\nabla \mathcal{J}(\bm{\theta}))$ 
		       \STATE $\mathcal{G}_i \leftarrow \{\bm{f}(\bm{\pi}_{i}^{1}),\dots,\bm{f}(\bm{\pi}_{i}^{n})\} \quad \forall i$
                  \STATE $\mathcal{F}_i \leftarrow {\rm{MPO}}(\tilde{\mathcal{F}}_i \cup \tilde{\mathcal{G}}_i) \quad \forall i$
             \ENDFOR
		\ENDFOR	
		\STATE \textbf{Output:} The model parameter $\bm{\theta}$
 	\end{algorithmic}
\end{algorithm}

In the inference phase, for $N$ given weights and diversity factors, the well-trained model is used to sequentially solve $N$ corresponding subproblems, as shown in Figure \ref{fig:frame}. Moreover, instance augmentation \cite{lin22} can be also brought into our MPO for each subproblem, i.e., the \emph{points} of all sampled solutions from an instance and its augmented instances are included in $\mathcal{G}^i$. Our NHDE can achieve desirable performance by using only parts of instance augmentation (see Appendix D), since it can already deliver more diverse solutions.

\section{Experiments}

\textbf{Problems.} We evaluate the proposed NHDE on three typical MOCO problems that are commonly studied in the neural MOCO literature \cite{lik21,lin22,zha22}, namely the multi-objective traveling salesman problem (MOTSP) \cite{lus10}, multi-objective capacitated vehicle routing problem (MOCVRP) \cite{zaj21}, and multi-objective knapsack problem (MOKP) \cite{ish15}. More specifically, we solve bi-objective TSP (Bi-TSP), tri-objective TSP (Tri-TSP), the bi-objective CVRP (Bi-CVRP) and the bi-objective KP (Bi-KP). For the $M$-objective TSP with $n$ nodes, each node has $M$ sets of 2-dimensional coordinates, where the $m$-th objective value of the solution is calculated with respect to the $m$-th coordinates. Bi-CVRP consists of $n$ customer nodes and a depot node, with each node featured by a 2-dimensional coordinate and each customer node associated with a demand. Following the literature, we consider two conflicting objectives in Bi-CVRP, i.e., the total tour length and the makespan (that is the length of the longest route). Bi-KP is defined by $n$ items, with each taking a weight and two separate values. The $m$-th objective is to maximize the sum of the $m$-th values but not exceed the capacity. Three sizes of these problems are considered, i.e., $n$=20/50/100 for MOTSP and MOCVRP, and $n$=50/100/200 for MOKP. The coordinates, demands, and values are uniformly sampled from $[0,1]^2$, $\{1,\dots,9\}$, and $[0,1]$, respectively. The vehicle capacity is set to 30/40/50 for MOCVRP20/50/100. The knapsack capacity is set to 12.5/25/25 for MOKP50/100/200. 

\textbf{Hyperparameters.} We train NHDE-P with 200 epochs, each containing 5,000 randomly generated instances. We use batch size $B\!=\!64$ and the Adam \cite{kin15} optimizer with learning rate $10^{-4}$ ($10^{-5}$ for MOKP) and weight decay $10^{-6}$. 
During training, $N'\!=\!20$ weights are sampled for each instance. During inference, we generate $N\!=\!40$ and $N\!=\!210$ uniformly distributed weights for $M\!=\!2$ and $M\!=\!3$, respectively, which are then shuffled so as to counteract biases. The diversity factors linearly shift through the $N$ subproblems from (1,0) to (0,1), which implies a gradual focus from achieving convergence (scalar objective) with a few solutions to ensuring comprehensive performance with a multitude of solutions. We set $K\!=\!20$ and $J\!=\!200$. See Appendix F for the settings of NHDE-M.

\textbf{Baselines.} We compare NHDE with two classes of state-of-the-art methods. (1) The neural methods, including \textbf{PMOCO} \cite{lin22}, \textbf{MDRL} \cite{zha22}, and DRL-based multiobjective optimization algorithm (\textbf{DRL-MOA}) \cite{lik21}, all with POMO as the backbone for single-objective CO subproblems. Our NHDE-P and NHDE-M each train a unified model with the same gradient steps as PMOCO and MDRL, respectively, while DRL-MOA trains 101 (105) models for $M=2$ ($M=3$) with more gradient steps, i.e., the first model with 200 epochs and the remaining models with 5-epoch per model via parameter transfer. (2) The non-learnable methods, including the state-of-the-art MOEA and strong heuristics. Particularly, \textbf{PPLS/D-C} \cite{shi22} is a specialized MOEA for MOCO with local search techniques, including a 2-opt heuristic for MOTSP and MOCVRP, and a greedy transformation heuristic \cite{ish15} for MOKP, implemented in Python. In addition, LKH \cite{hel00,tin18} and dynamic programming (DP), are employed to solve the weighted-sum (WS) based subproblems for MOTSP and MOKP, denoted as \textbf{WS-LKH} and \textbf{WS-DP}, respectively. All the methods use WS scalarization for fair comparisons. All the methods are tested with an RTX 3090 GPU and an Intel Xeon 4216 CPU. Our code is publicly available\footnote{\url{https://github.com/bill-cjb/NHDE}}.

\textbf{Metrics.} We use hypervolume (HV) and the number of non-dominated solutions ($|$NDS$|$). A higher HV means better overall performance in terms of convergence and diversity, while $|$NDS$|$ reflects the diversity when HVs are close. The average HV, gaps with respect to NHDE, and total running time for 200 random test instances are reported. The best (second-best) and its statistically insignificant results at 1\% significance level of a Wilcoxon rank-sum test are highlighted in \textbf{bold} (\underline{underline}).

\subsection{Main results}
All results of NHDE-P and the baselines are displayed in Table \ref{tab:main}. Given the same number of weights (wt.), NHDE-P significantly surpasses PMOCO for all problems and sizes in terms of HV and $|$NDS$|$, which indicates that NHDE-P has the potential to discover diverse and high-quality solutions.  
When instance augmentation (aug.) is equipped, NHDE-P achieves the smallest gap among the methods in most cases, except Bi-TSP100 and Bi-CVRP100 where WS-LKH and DRL-MOA perform better. However, WS-LKH consumes much longer runtime than NHDE-P due to iterative search (2.7 hours vs 5.6 minutes), and DRL-MOA costs much more training overhead to prepare multiple models for respective weights. Besides, another reason why NHDE-P is inferior to DRL-MOA on Bi-CVRP100 might be that the hypernetwork (inherited from PMOCO) could be hard to cope with the objectives with imbalanced scales. 
Considering this drawback of PMOCO, we also apply our method to MDRL, i.e., NHDE-M, and demonstrate that NHDE-M outperforms DRL-MOA on Bi-CVRP100 in Table \ref{tab:main-m}. Also, NHDE-M outperforms MDRL in all cases. More results of NHDE-M are given in Appendix G.

\begin{table}[!t]
  \centering
  \caption{Results of NHDE-P on 200 random instances for MOCO problems.}
  \resizebox{\textwidth}{!}{
  	\addtolength{\tabcolsep}{-4pt}
    \begin{tabular}{l|cccc|cccc|cccc}
    \toprule
          & \multicolumn{4}{c|}{Bi-TSP20} & \multicolumn{4}{c|}{Bi-TSP50} & \multicolumn{4}{c}{Bi-TSP100} \\
    Method & HV$\uparrow$    & $|$NDS$|$$\uparrow$ & Gap$\downarrow$   & Time  & HV$\uparrow$    & $|$NDS$|$$\uparrow$ & Gap$\downarrow$   & Time  & HV$\uparrow$    & $|$NDS$|$$\uparrow$ & Gap$\downarrow$   & Time \\
    \midrule
    WS-LKH (40 wt.) & 0.6266  & 14    & 0.46\% & 4.1m  & \underline{0.6402}  & 29    & \underline{0.42\%} & 42m   & \textbf{0.7072} & 37    & \textbf{-0.31\%} & 2.7h \\
    PPLS/D-C (200 iter.) & 0.6256  & 71    & 0.62\% & 26m   & 0.6282  & 213   & 2.29\% & 2.8h  & 0.6844  & 373   & 2.92\% & 11h \\
    DRL-MOA (101 models) & 0.6257  & 23    & 0.60\% & 6s    & 0.6360  & 57    & 1.07\% & 9s    & 0.6970  & 70    & 1.13\% & 21s \\
    \midrule
    PMOCO (40 wt.) & 0.6258  & 17    & 0.59\% & 4s    & 0.6331  & 31    & 1.52\% & 5s    & 0.6938  & 36    & 1.59\% & 8s \\
    PMOCO (600 wt.) & 0.6267  & 23    & 0.44\% & 27s   & 0.6361  & 68    & 1.06\% & 53s   & 0.6978  & 131   & 1.02\% & 2.1m \\
    NHDE-P (40 wt.) & \underline{0.6286}  & 56    & \underline{0.14\%} & 19s   & 0.6388  & 127   & 0.64\% & 53s   & 0.7005  & 193   & 0.64\% & 1.9m \\
    \midrule
    PMOCO (40 wt. aug.) & 0.6266  & 17    & 0.46\% & 23s   & 0.6377  & 32    & 0.81\% & 1.6m  & 0.6993  & 37    & 0.81\% & 3.0m \\
    PMOCO (100 wt. aug.) & 0.6270  & 20    & 0.40\% & 1.4m  & 0.6395  & 53    & 0.53\% & 3.8m  & 0.7016  & 76    & 0.48\% & 15m \\
    NHDE-P (40 wt. aug.) & \textbf{0.6295} & 81    & \textbf{0.00\%} & 1.5m  & \textbf{0.6429} & 269   & \textbf{0.00\%} & 2.5m  & \underline{0.7050}  & 343   & \underline{0.00\%} & 5.6m \\
    \midrule
          & \multicolumn{4}{c|}{Bi-CVRP20} & \multicolumn{4}{c|}{Bi-CVRP50} & \multicolumn{4}{c}{Bi-CVRP100} \\
    Method & HV$\uparrow$    & $|$NDS$|$$\uparrow$ & Gap$\downarrow$   & Time  & HV$\uparrow$    & $|$NDS$|$$\uparrow$ & Gap$\downarrow$   & Time  & HV$\uparrow$    & $|$NDS$|$$\uparrow$ & Gap$\downarrow$   & Time \\
    \midrule
    PPLS/D-C (200 iter.) & 0.4283  & 14    & 0.46\% & 1.3h  & 0.4007  & 17    & 2.15\% & 9.7h  & 0.3946  & 20    & 1.13\% & 38h \\
    DRL-MOA (101 models) & 0.4287  & 7     & 0.37\% & 10s   & 0.4076  & 10    & 0.46\% & 12s   & \textbf{0.4055} & 12    & \textbf{-1.60\%} & 33s \\
    \midrule
    PMOCO (40 wt.) & 0.4266  & 6     & 0.86\% & 4s    & 0.4035  & 7     & 1.47\% & 7s    & 0.3912  & 6     & 1.98\% & 12s \\
    PMOCO (300 wt.) & 0.4268  & 7     & 0.81\% & 20s   & 0.4039  & 9     & 1.37\% & 35s   & 0.3914  & 8     & 1.93\% & 1.2m \\
    NHDE-P (40 wt.) & 0.4284  & 12    & 0.44\% & 18s   & 0.4062  & 14    & 0.81\% & 36s   & 0.3933  & 10    & 1.45\% & 1.1m \\
    \midrule
    PMOCO (40 wt. aug.) & 0.4292  & 6     & 0.26\% & 8s    & 0.4078  & 7     & 0.42\% & 15s   & 0.3968  & 7     & 0.58\% & 1.1m \\
    PMOCO (300 wt. aug.) & \underline{0.4294}  & 9     & \underline{0.21\%} & 1.0m  & \underline{0.4081}  & 10    & \underline{0.34\%} & 1.8m  & 0.3969  & 9     & 0.55\% & 7.0m \\
    NHDE-P (40 wt. aug.) & \textbf{0.4303} & 21    & \textbf{0.00\%} & 1.2m  & \textbf{0.4095} & 22    & \textbf{0.00\%} & 1.5m  & \underline{0.3991}  & 16    & \underline{0.00\%} & 2.4m \\
    \midrule
          & \multicolumn{4}{c|}{Bi-KP50}  & \multicolumn{4}{c|}{Bi-KP100} & \multicolumn{4}{c}{Bi-KP200} \\
    Method & HV$\uparrow$    & $|$NDS$|$$\uparrow$ & Gap$\downarrow$   & Time  & HV$\uparrow$    & $|$NDS$|$$\uparrow$ & Gap$\downarrow$   & Time  & HV$\uparrow$    & $|$NDS$|$$\uparrow$ & Gap$\downarrow$   & Time \\
    \midrule
    WS-DP (40 wt.) & \underline{0.3560}  & 10    & \underline{0.11\%} & 9.6m  & 0.4529  & 16    & 0.26\% & 1.3h  & 0.3598  & 23    & 0.39\% & 3.8h \\
    PPLS/D-C (200 iter.) & 0.3528  & 13    & 1.01\% & 18m   & 0.4480  & 19    & 1.34\% & 47m   & 0.3541  & 20    & 1.97\% & 1.5h \\
    DRL-MOA (101 models) & 0.3559  & 21    & 0.14\% & 9s    & \underline{0.4531}  & 38    & \underline{0.22\%} & 18s   & \underline{0.3601}  & 48    & \underline{0.30\%} & 1.0m \\
    \midrule
    PMOCO (40 wt.) & 0.3550  & 14    & 0.39\% & 6s    & 0.4518  & 22    & 0.51\% & 9s    & 0.3590  & 28    & 0.61\% & 25s \\
    PMOCO (300 wt.) & 0.3552  & 17    & 0.34\% & 29s   & 0.4524  & 31    & 0.37\% & 1.1m  & 0.3597  & 46    & 0.42\% & 3.0m \\
    NHDE-P (40 wt.) & \textbf{0.3564} & 30    & \textbf{0.00\%} & 29s   & \textbf{0.4541} & 83    & \textbf{0.00\%} & 1.0m  & \textbf{0.3612} & 243   & \textbf{0.00\%} & 2.7m \\
    \midrule
          & \multicolumn{4}{c|}{Tri-TSP20} & \multicolumn{4}{c|}{Tri-TSP50} & \multicolumn{4}{c}{Tri-TSP100} \\
    Method & HV$\uparrow$    & $|$NDS$|$$\uparrow$ & Gap$\downarrow$   & Time  & HV$\uparrow$    & $|$NDS$|$$\uparrow$ & Gap$\downarrow$   & Time  & HV$\uparrow$    & $|$NDS$|$$\uparrow$ & Gap$\downarrow$   & Time \\
    \midrule
    WS-LKH (210 wt.) & 0.4727  & 78    & 0.82\% & 23m   & 0.4501  & 189   & 1.90\% & 3.5h  & \underline{0.5165}  & 209   & \underline{0.84\%} & 12h \\
    PPLS/D-C (200 iter.) & 0.4698  & 876   & 1.43\% & 1.4h  & 0.4174  & 3727  & 9.02\% & 3.9h  & 0.4376  & 8105  & 15.99\% & 14h \\
    DRL-MOA (105 models) & 0.4675  & 72    & 1.91\% & 5s    & 0.4285  & 98    & 6.61\% & 9s    & 0.4850  & 101   & 6.89\% & 19s \\
    \midrule
    PMOCO (210 wt.) & 0.4714  & 113   & 1.09\% & 11s   & 0.4381  & 198   & 4.51\% & 18s   & 0.4946  & 207   & 5.05\% & 39s \\
    PMOCO (3003 wt.) & 0.4741  & 264   & 0.52\% & 2.3m  & 0.4484  & 1339  & 2.27\% & 4.6m  & 0.5087  & 2330  & 2.34\% & 10m  \\
    NHDE-P (210 wt.) & \underline{0.4758}  & 675   & \underline{0.17\%} & 1.2m  & \underline{0.4506}  & 2547  & \underline{1.79\%} & 4.4m  & 0.5111  & 4984  & 1.88\% & 10m \\
    \midrule
    PMOCO (210 wt. aug.) & 0.4727  & 104   & 0.82\% & 21m   & 0.4471  & 201   & 2.55\% & 1.1h  & 0.5044  & 209   & 3.17\% & 4.2h \\
    PMOCO (153 wt. aug.) & 0.4722  & 89    & 0.92\% & 15m   & 0.4447  & 150   & 3.07\% & 47m   & 0.5009  & 153   & 3.84\% & 3.1h \\
    NHDE-P (210 wt. aug.) & \textbf{0.4766} & 527   & \textbf{0.00\%} & 14m   & \textbf{0.4588} & 9047  & \textbf{0.00\%} & 30m   & \textbf{0.5209} & 16999  & \textbf{0.00\%} & 1.5h \\
    \bottomrule
    \end{tabular}%
    }
  \label{tab:main}%
\end{table}%

\begin{table}[!t]
  \centering
  \caption{Results of NHDE-M on 200 random instances for MOCO problems.}
  \resizebox{\textwidth}{!}{
  	\addtolength{\tabcolsep}{-4pt}
    \begin{tabular}{l|cccc|cccc|cccc}
    \toprule
          & \multicolumn{4}{c|}{Bi-TSP20} & \multicolumn{4}{c|}{Bi-TSP50} & \multicolumn{4}{c}{Bi-TSP100} \\
    Method & HV$\uparrow$    & $|$NDS$|$$\uparrow$ & Gap$\downarrow$   & Time  & HV$\uparrow$    & $|$NDS$|$$\uparrow$ & Gap$\downarrow$   & Time  & HV$\uparrow$    & $|$NDS$|$$\uparrow$ & Gap$\downarrow$   & Time \\
    \midrule
    WS-LKH (40 wt.) & 0.6266  & 14    & 0.46\% & 4.1m  & \underline{0.6402}  & 29    & \underline{0.42\%} & 42m   & \textbf{0.7072} & 37    & \textbf{-0.33\%} & 2.7h \\
    PPLS/D-C (200 iter.) & 0.6256  & 71    & 0.62\% & 26m   & 0.6282  & 213   & 2.29\% & 2.8h  & 0.6844  & 373   & 2.91\% & 11h \\
    DRL-MOA (101 models) & 0.6257  & 23    & 0.60\% & 6s    & 0.6360  & 57    & 1.07\% & 9s    & 0.6970  & 70    & 1.12\% & 21s \\
    \midrule
    MDRL (40 wt.) & 0.6264  & 20    & 0.49\% & 2s    & 0.6342  & 33    & 1.35\% & 3s    & 0.6940  & 36    & 1.55\% & 8s \\
    NHDE-M (40 wt.) & \underline{0.6287}  & 58    & \underline{0.13\%} & 20s   & 0.6393  & 132   & 0.56\% & 57s   & 0.7008  & 195   & 0.58\% & 2.0m \\
    \midrule
    MDRL (40 wt. aug.) & 0.6267  & 18    & 0.44\% & 21s   & 0.6384  & 34    & 0.70\% & 1.5m  & 0.6995  & 38    & 0.77\% & 3.3m \\
    NHDE-M (40 wt. aug.) & \textbf{0.6295} & 81    & \textbf{0.00\%} & 1.5m  & \textbf{0.6429} & 273   & \textbf{0.00\%} & 2.6m  & \underline{0.7049}  & 339   & \underline{0.00\%} & 5.5m \\
    \midrule
          & \multicolumn{4}{c|}{Bi-CVRP20} & \multicolumn{4}{c|}{Bi-CVRP50} & \multicolumn{4}{c}{Bi-CVRP100} \\
    Method & HV$\uparrow$    & $|$NDS$|$$\uparrow$ & Gap$\downarrow$   & Time  & HV$\uparrow$    & $|$NDS$|$$\uparrow$ & Gap$\downarrow$   & Time  & HV$\uparrow$    & $|$NDS$|$$\uparrow$ & Gap$\downarrow$   & Time \\
    \midrule
    PPLS/D-C (200 iter.) & 0.4287  & 15    & 0.42\% & 1.6h  & 0.4007  & 17    & 2.34\% & 9.7h  & 0.3946  & 20    & 3.14\% & 38h \\
    DRL-MOA (101 models) & 0.4287  & 7     & 0.42\% & 10s   & 0.4076  & 10    & 0.66\% & 12s   & \underline{0.4055}  & 12    & \underline{0.47\%} & 33s \\
    \midrule
    MDRL (40 wt.) & 0.4284  & 9     & 0.49\% & 3s    & 0.4057  & 5     & 1.12\% & 5s    & 0.4015  & 0     & 1.45\% & 10s \\
    NHDE-M (40 wt.) & \underline{0.4296}  & 16    & \underline{0.21\%} & 23s   & \underline{0.4086}  & 20    & \underline{0.41\%} & 47s   & 0.4053  & 18    & 0.52\% & 1.4m \\
    \midrule
    MDRL (40 wt. aug.) & 0.4293  & 9     & 0.28\% & 5s    & 0.4073  & 11    & 0.73\% & 16s   & 0.4040  & 11    & 0.83\% & 1.0m \\
    NHDE-M (40 wt. aug.) & \textbf{0.4305} & 24    & \textbf{0.00\%} & 1.2m  & \textbf{0.4103} & 29    & \textbf{0.00\%} & 1.6m  & \textbf{0.4074} & 26    & \textbf{0.00\%} & 2.7m \\
    \midrule
          & \multicolumn{4}{c|}{Bi-KP50}  & \multicolumn{4}{c|}{Bi-KP100} & \multicolumn{4}{c}{Bi-KP200} \\
    Method & HV$\uparrow$    & $|$NDS$|$$\uparrow$ & Gap$\downarrow$   & Time  & HV$\uparrow$    & $|$NDS$|$$\uparrow$ & Gap$\downarrow$   & Time  & HV$\uparrow$    & $|$NDS$|$$\uparrow$ & Gap$\downarrow$   & Time \\
    \midrule
    WS-DP (40 wt.) & \underline{0.3560}  & 10    & \underline{0.17\%} & 9.6m  & 0.4529  & 16    & 0.29\% & 1.3h  & 0.3598  & 23    & 0.30\% & 3.8h \\
    PPLS/D-C (200 iter.) & 0.3528  & 13    & 1.07\% & 18m   & 0.4480  & 19    & 1.37\% & 47m   & 0.3541  & 20    & 1.88\% & 1.5h \\
    DRL-MOA (101 models) & 0.3559  & 21    & 0.20\% & 9s    & \underline{0.4531}  & 38    & \underline{0.24\%} & 18s   & \underline{0.3601}  & 48    & \underline{0.22\%} & 1.0m \\
    \midrule
    MDRL (40 wt.) & 0.3559  & 17    & 0.20\% & 4s    & 0.4528  & 25    & 0.31\% & 8s    & 0.3594  & 31    & 0.42\% & 24s \\
    NHDE-M (40 wt.) & \textbf{0.3566} & 41    & \textbf{0.00\%} & 31s   & \textbf{0.4542} & 93    & \textbf{0.00\%} & 1.0m  & \textbf{0.3609} & 160   & \textbf{0.00\%} & 2.8m \\
    \midrule
          & \multicolumn{4}{c|}{Tri-TSP20} & \multicolumn{4}{c|}{Tri-TSP50} & \multicolumn{4}{c}{Tri-TSP100} \\
    Method & HV$\uparrow$    & $|$NDS$|$$\uparrow$ & Gap$\downarrow$   & Time  & HV$\uparrow$    & $|$NDS$|$$\uparrow$ & Gap$\downarrow$   & Time  & HV$\uparrow$    & $|$NDS$|$$\uparrow$ & Gap$\downarrow$   & Time \\
    \midrule
    WS-LKH (210 wt.) & 0.4727  & 78    & 0.82\% & 23m   & 0.4501  & 189   & 2.00\% & 3.5h  & \textbf{0.5165} & 209   & \textbf{-0.92\%} & 12h \\
    PPLS/D-C (200 iter.) & 0.4698  & 876   & 1.43\% & 1.4h  & 0.4174  & 3727  & 9.12\% & 3.9h  & 0.4376  & 8105  & 14.50\% & 14h \\
    DRL-MOA (105 models) & 0.4675  & 72    & 1.91\% & 5s    & 0.4285  & 98    & 6.71\% & 9s    & 0.4850  & 101   & 5.24\% & 19s \\
    \midrule
    MDRL (210 wt.) & 0.4723  & 126   & 0.90\% & 14s   & 0.4388  & 199   & 4.46\% & 20s   & 0.4956  & 207   & 3.17\% & 40s \\
    NHDE-M (210 wt.) & \underline{0.4763}  & 783   & \underline{0.06\%} & 1.4m  & \underline{0.4512}  & 2636  & \underline{1.76\%} & 4.7m  & 0.4997  & 4056  & 2.36\% & 11m \\
    \midrule
    MDRL (210 wt. aug.) & 0.4727  & 107   & 0.82\% & 14m   & 0.4473  & 202   & 2.61\% & 53m   & 0.5056  & 209   & 1.21\% & 4.3h \\
    NHDE-M (210 wt. aug.) & \textbf{0.4766} & 748   & \textbf{0.00\%} & 13m   & \textbf{0.4593} & 10850  & \textbf{0.00\%} & 30m   & \underline{0.5118}  & 13216  & \underline{0.00\%} & 1.5h \\
    \bottomrule
    \end{tabular}%
    }
  \label{tab:main-m}%
\end{table}%

Regarding the inference efficiency, NHDE-P generally takes (tolerably) more runtime than the other learning based methods. To be fair, we further enhance the state-of-the-art PMOCO by adjusting the number of the weights, so that it takes similar or even more runtime compared with NHDE-P. The results in the last two lines for each problem shows that NHDE-P still attains the smaller gaps. We also observe that $|$NDS$|$ hardly grows along with the increase of number of weights for PMOCO, since numerous solutions to different subproblems could be repetitive. In contrast, NHDE-P is able to produce more diverse solutions with much fewer weights.

\subsection{Generalization study}

\begin{figure}[!t]
	\centering
	\subfigure[]{
		\centering
		\includegraphics[width=0.31\textwidth]{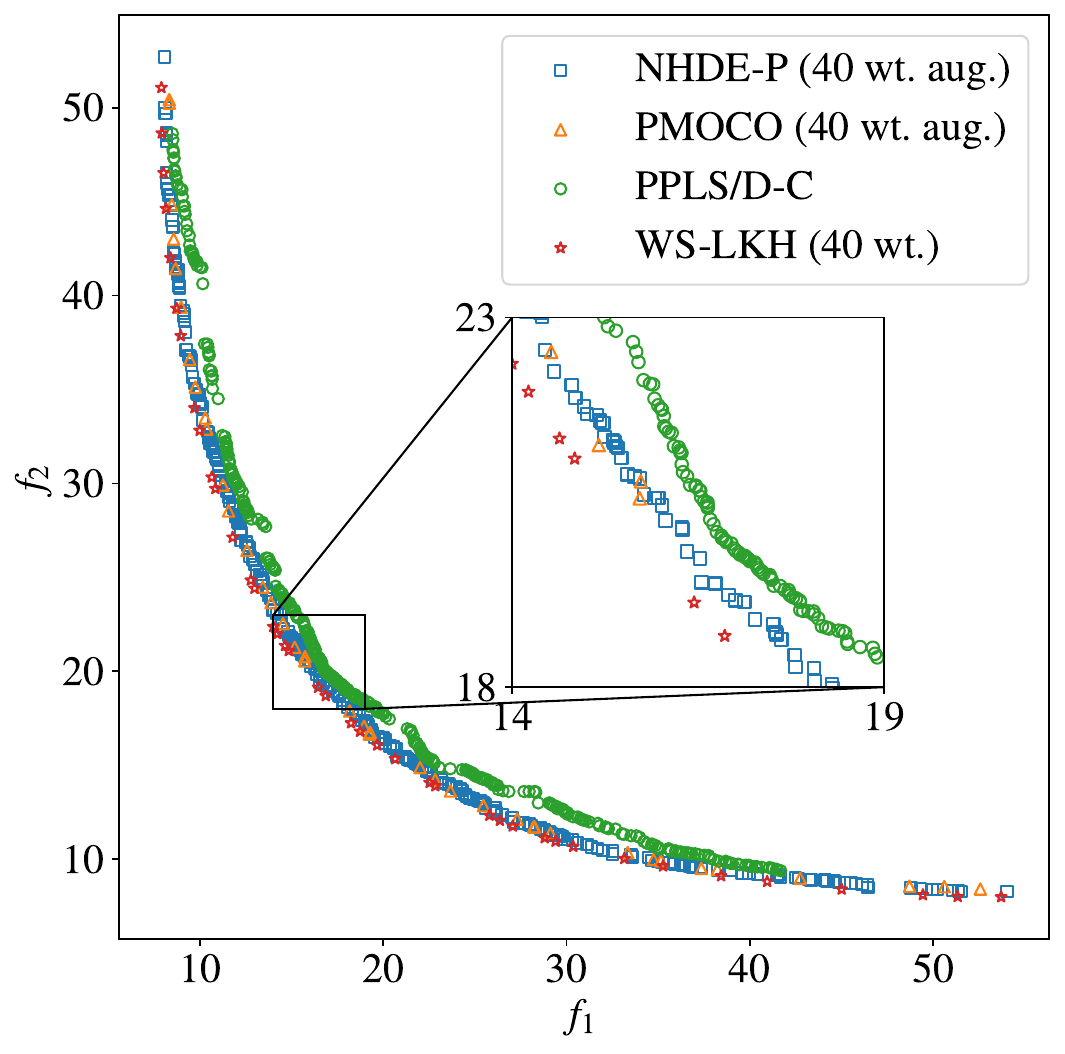}
		\label{fig:kro100}
	}
	\subfigure[]{
		\centering
		\includegraphics[width=0.31\textwidth]{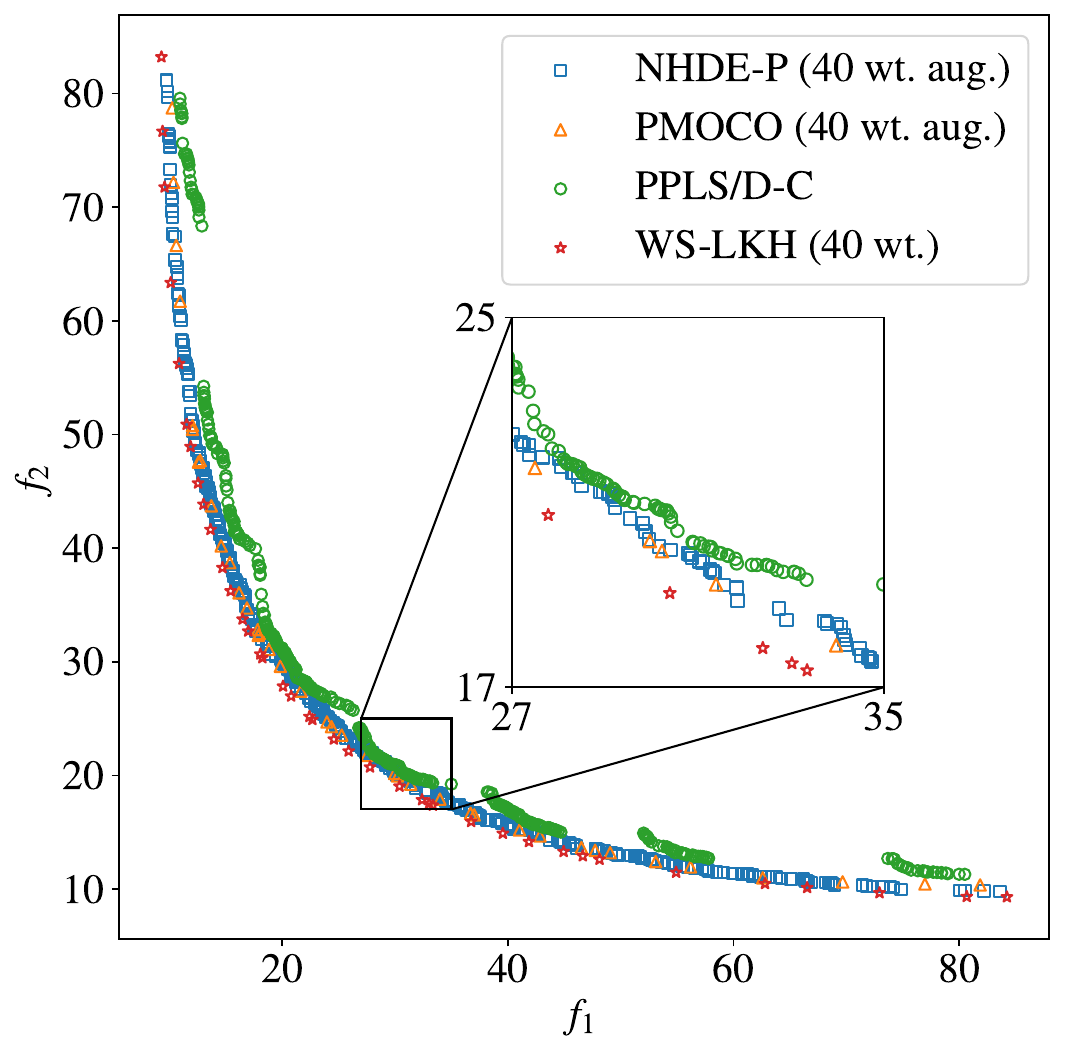}
		\label{fig:kro150}
	}
	\subfigure[]{
		\centering
		\includegraphics[width=0.31\textwidth]{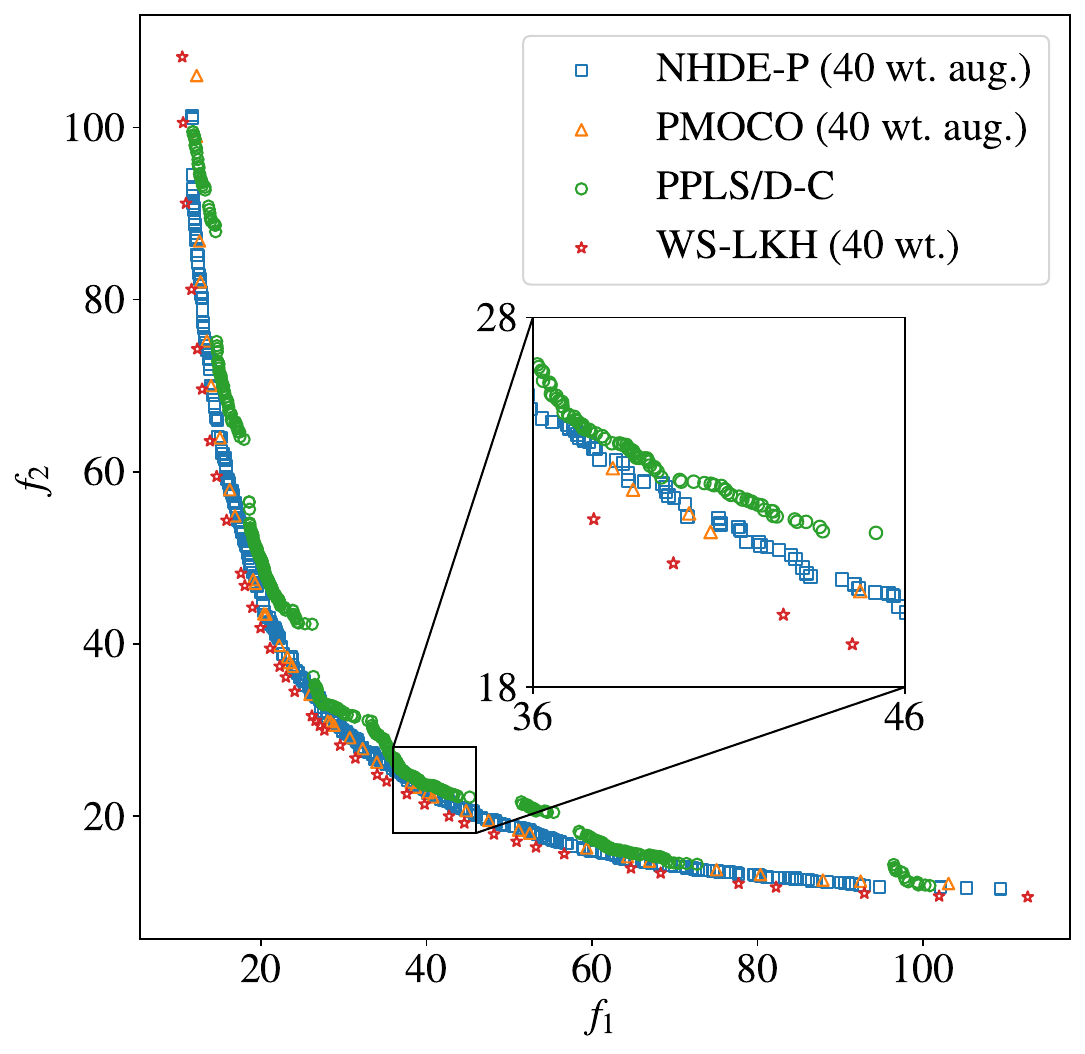}
		\label{fig:kro200}
	}
	\caption{Pareto fronts of benchmark instances. (a) KroAB100. (b) KroAB150. (c) KroAB200.}
	\label{fig:kro}
\end{figure}

To assess the generalization capability of NHDE-P, we compare the trained models (from neural methods) for Bi-TSP100 and the other baselines on 200 random Bi-TSP instances with larger sizes, i.e., Bi-TSP150/200. Three commonly used benchmark instances developed from TSPLIB \cite{rei91}, i.e., KroAB100, KroAB150, and KroAB200, are also tested. The comparison results and Pareto fronts are demonstrated in Table \ref{tab:gen} and Figure \ref{fig:kro}, respectively. As shown, NHDE-P outperforms the state-of-the-art MOEA (i.e., PPLS/D-C) and other neural methods significantly, in terms of HV and $|$NDS$|$, which means a superior generalization capability. The figure again verifies that NHDE-P generates a more extended Pareto front than PMOCO, showing a better diversity. Although PPLS/D-C finds a large number of solutions, they are inferior in terms of the optimality, with a biased distribution (e.g., crowded in certain regions). In contrast, NHDE-P generates a more well-distributed Pareto front with stronger convergence. More results on generalization are given in Appendix H.

\begin{table}[!t]
  \centering
  \caption{Results on 200 random instances for larger-scale problems.}
    \resizebox{\textwidth}{!}{
    \begin{tabular}{l|cccc|cccc}
    \toprule
          & \multicolumn{4}{c|}{Bi-TSP150} & \multicolumn{4}{c}{Bi-TSP200} \\
    Method & HV$\uparrow$    & $|$NDS$|$$\uparrow$ & Gap$\downarrow$   & Time  & HV$\uparrow$    & $|$NDS$|$$\uparrow$ & Gap$\downarrow$   & Time \\
    \midrule
    WS-LKH (40 wt.) & \textbf{0.7075} & 39    & \textbf{-0.90\%} & 5.3h  & \textbf{0.7435} & 40    & \textbf{-1.52\%} & 8.5h \\
    PPLS/D-C (200 iter.) & 0.6784  & 473   & 3.25\% & 21h   & 0.7106  & 512   & 2.98\% & 32h \\
    DRL-MOA (101 models) & 0.6901  & 73    & 1.58\% & 45s   & 0.7219  & 75    & 1.43\% & 87s \\
    \midrule
    PMOCO (40 wt.) & 0.6891  & 37    & 1.73\% & 22s   & 0.7215  & 38    & 1.49\% & 41s \\
    PMOCO (400 wt.) & 0.6938  & 160   & 1.06\% & 3.7m  & 0.7259  & 186   & 0.89\% & 6.8m \\
    NHDE-P (40 wt.) & 0.6964  & 231   & 0.68\% & 3.0m  & 0.7280  & 259   & 0.60\% & 4.3m \\
    \midrule
    PMOCO (40 wt. aug.) & 0.6944  & 38    & 0.97\% & 20m   & 0.7264  & 39    & 0.82\% & 40m \\
    NHDE-P (40 wt. aug.) & \underline{0.7012}  & 372   & \underline{0.00\%} & 14m   & \underline{0.7324}  & 384   & \underline{0.00\%} & 26m \\
    \bottomrule
    \end{tabular}%
    }
  \label{tab:gen}%
\end{table}%

\subsection{Ablation Study}

To analyze the effect of the indicator-enhanced DRL, we compare NHDE-P with decomposition-based DRL without indicator (NHDE w/o I) and indicator-based DRL without decomposition (NHDE w/o D). To verify the valid design of HGA, three attention-based variants, which are formed by removing \emph{point-to-node} attention (NHDE w/o P2N), removing \emph{node-to-point} attention (NHDE w/o N2P), and adding \emph{point-to-point} attention (NHDE w P2P), are involved for comparison. To assess the impact of MPO, NHDE w/o MPO is also evaluated. More details about these variants are presented in Appendix I. As seen from Figure \ref{fig:abla}. The performance of NHDE-P is significantly impaired when any of the components is ablated. Instead, NHDE w P2P degrades a bit, which reveals that the extra \emph{point-to-point} attention may bring noises into the model. Moreover, we evaluate the effectiveness of the efficient update of MPO and partial instance augmentation in Figure \ref{fig:effi}, which shows that either of them saliently diminishes solving time with only a little sacrifice of the performance.

\begin{figure}[!t]
	\centering
	\subfigure[]{
		\centering
		\includegraphics[width=0.72\textwidth]{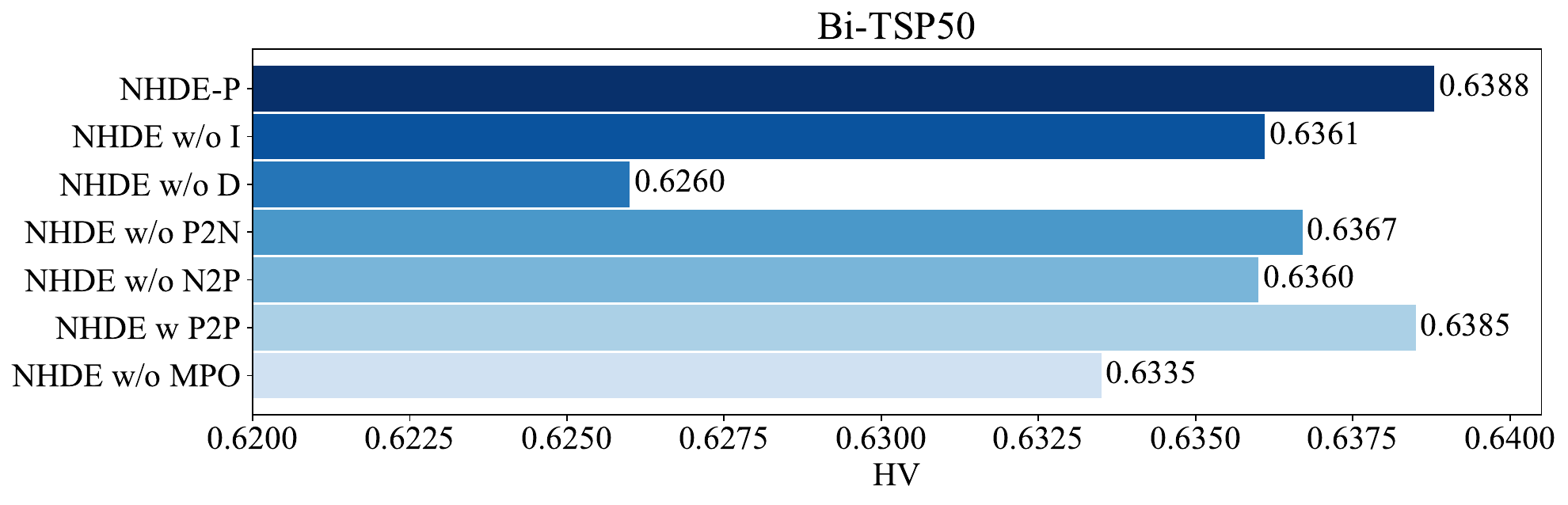}
		\label{fig:abla}
	}
	\subfigure[]{
		\centering
		\includegraphics[width=0.24\textwidth]{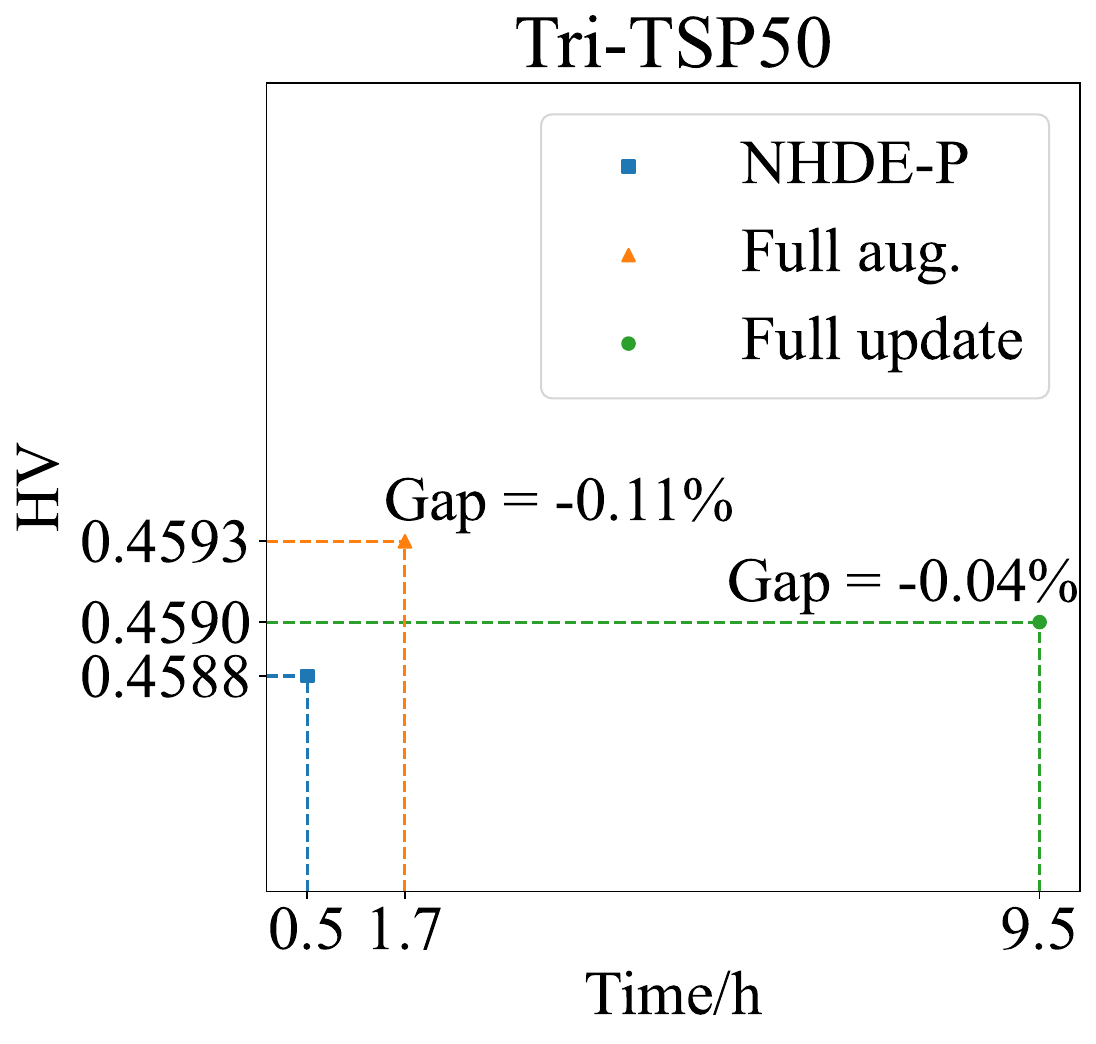}
		\label{fig:effi}
	}
	\caption{Ablation study. (a) Effects of indicator-enhanced DRL, HGA, and MPO. (b) Effects of the efficient update of MPO and partial instance augmentation.}
	\label{fig:effe}
\end{figure}

\section{Conclusion}

This paper proposes a novel NHDE for MOCO problems. NHDE impedes repetitive solutions from different subproblems via indicator-enhanced DRL with a HGA model, and digs more solutions in the neighborhood of each subproblem with an MPO strategy. Our generic NHDE can be deployed to different neural MOCO methods. The experimental results on three classic MOCO problems showed the superiority of NHDE, especially with regard to the diversity of Pareto set. A limitation is that the HV calculation would expend additional computational time, which might hinder the scalability of NHDE for solving much larger problems with many objectives. In the future, we will explore alternative schemes like the HV approximations \cite{boe23,sha22} to further promote the training efficiency of NHDE, and we also intend to apply it to tackle real-world MOCO problems.

\section*{Acknowledgments and disclosure of funding}

This work is supported by the National Natural Science Foundation of China (62072483), and the Guangdong Basic and Applied Basic Research Foundation (2022A1515011690, 2021A1515012298).


{
\small
\bibliographystyle{unsrtnat}
\bibliography{ref}
}

\clearpage

\appendix
\setcounter{page}{1} 
\appendix

\vbox{
\hrule height 4pt
\vskip 0.25in
\vskip -\parskip%
\centering
{\LARGE\bf Neural Multi-Objective Combinatorial Optimization with Diversity Enhancement (Appendix)}
\vskip 0.29in
\vskip -\parskip
\hrule height 1pt
}

\section{Reference point and hypervolume ratio}

The normalized hypervolume (HV) ratio is calculated as ${\rm{HV}}'_{\bm{r}}(\mathcal{F})={\rm{HV}}_{\bm{r}}(\mathcal{F})/\prod_{i=1}^{M}|r_i-z_i|$, where $\bm{r}$ is a reference point satisfying $r_i>\max\{f_i(\bm{x})|\bm{f}(\bm{x}) \in \mathcal{F}\}$ and $\bm{z}$ is an ideal point satisfying $z_i<\min\{f_i(\bm{x})|\bm{f}(\bm{x}) \in \mathcal{F}\}$\footnote{$r_i<\min\{f_i(\bm{x})|\bm{f}(\bm{x}) \in \mathcal{F}\}$ and $z_i>\max\{f_i(\bm{x})|\bm{f}(\bm{x}) \in \mathcal{F}\}$ for maximization problems, e.g., Bi-KP.}, $\forall i \in \{1,\dots, M\}$. The used $\bm{r}$ and $\bm{z}$ are given in Table \ref{tab:ref}.

\begin{table}[h]
  \centering
  \caption{Reference points and ideal points}
    \begin{tabular}{cccc}
    \toprule
    Problem & Size  & $\bm{r}$     & $\bm{z}$ \\
    \midrule
    \multirow{5}[2]{*}{Bi-TSP} & 20    & (20, 20) & (0, 0) \\
          & 50    & (35, 35) & (0, 0) \\
          & 100   & (65, 65) & (0, 0) \\
          & 150   & (85, 85) & (0, 0) \\
          & 200   & (115, 115) & (0, 0) \\
    \midrule
    \multirow{3}[2]{*}{Bi-CVRP} & 20    & (30, 4) & (0, 0) \\
          & 50    & (45, 4) & (0, 0) \\
          & 100   & (80, 4) & (0, 0) \\
    \midrule
    \multirow{3}[2]{*}{Bi-KP} & 50    & (5, 5) & (30, 30) \\
          & 100   & (20, 20) & (50, 50) \\
          & 200   & (30, 30) & (75, 75) \\
    \midrule
    \multirow{3}[2]{*}{Tri-TSP} & 20    & (20, 20, 20) & (0, 0) \\
          & 50    & (35, 35, 35) & (0, 0) \\
          & 100   & (65, 65, 65) & (0, 0) \\
    \bottomrule
    \end{tabular}%
  \label{tab:ref}%
\end{table}%

\section{Details of NHDE-P and NHDE-M}

\begin{figure}[!t]
    \centering
    \includegraphics[width=0.8\textwidth]{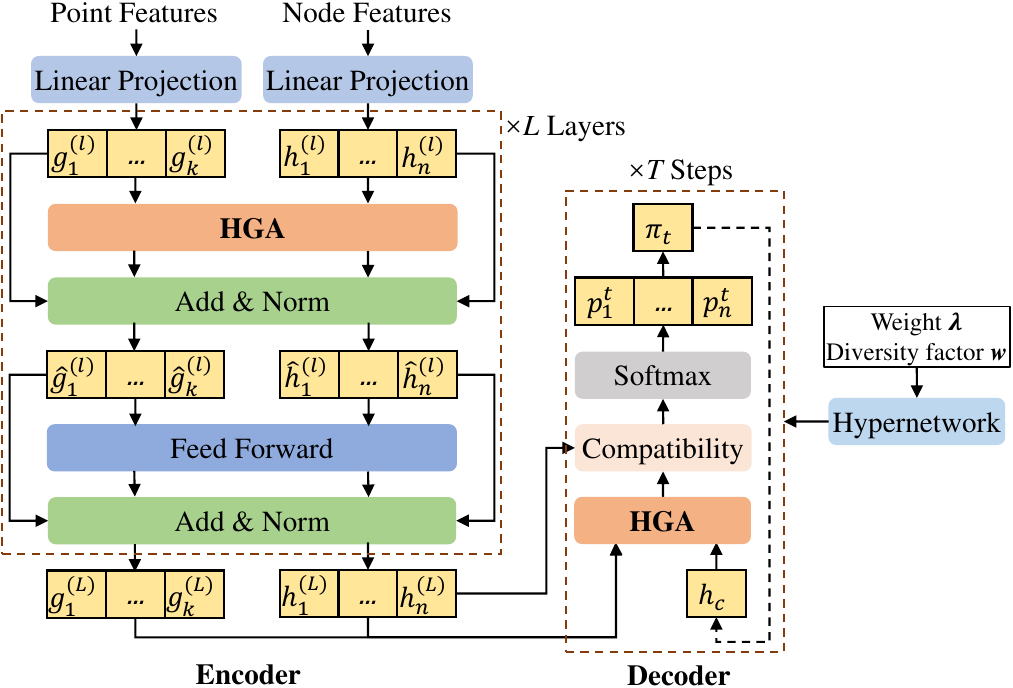}
    \caption{Hypernetwork-based heterogeneous graph attention (HGA) model of NHDE-P.}
    \label{fig:NHDE-P}
\end{figure}

NHDE-P, deploying NHDE to PMOCO \cite{lin22}, employs a hypernetwork to tackle the weight $\bm{\lambda}$ and diversity factor $\bm{w}$ for the corresponding subproblem. Specifically, according to the given $\bm{\lambda}$ and $\bm{w}$, the hypernetwork generates the decoder parameters of the heterogeneous graph attention (HGA) model $\bm{\theta}$, which is an encoder-decoder-styled architecture, i.e., $\bm{\theta}(\bm{\lambda},\bm{w}) = [\bm{\theta}_{\rm{en}},\bm{\theta}_{\rm{de}}(\bm{\lambda},\bm{w})]$, as shown in Figure \ref{fig:NHDE-P}. Following \cite{lin22}, the hypernetwork adopts a simple MLP model with two 256-dimensional hidden layers and ReLu activation. The MLP first maps an input with $M+2$ dimensions to a hidden embedding $\bm{h}(\bm{\lambda},\bm{w})$, which is then used to generate the decoder parameters by linear projection.

NHDE-M, deploying NHDE to MDRL \cite{zha22}, consists of three processes. In the meta-learning process, a meta-model $\bm{\theta}_{\rm{meta}}$, whose architecture is the same as the HGA model $\bm{\theta}$, is trained by sampling tasks from the whole task space. In the fine-tuning process, according to the given $\bm{\lambda}$ and $\bm{w}$, $\bm{\theta}_{\rm{meta}}$ is then fine-tuned using fine-tuning instances with a few gradient steps to derive the corresponding submodel. In the inference process, the submodel is used to solve the corresponding subproblem.

\section{Node features and context embedding}

The input dimensions of the node features vary with different problems. The inputs of the $M$-objective TSP are $n$ nodes with $2M$-dimensional features. The inputs of Bi-CVRP are $n$ customer nodes with 3-dimensional features and a depot node with 2-dimensional features. The inputs of Bi-KP are $n$ nodes with 3-dimensional features.

At step $t$ in the decoder, a context embedding $\bm{h}_c$ is used to calculated the probability of node selection. For MOTSP, $\bm{h}_c$ is defined as the concatenation of the graph embedding $\bar{\bm{h}}=\sum_{u=1}^n \bm{h}_u/n$, the embedding of the first node $\bm{h}_{\bm{\pi}_1}$, and the embedding of the last node $\bm{h}_{\bm{\pi}_{t-1}}$. For MOCVRP, $\bm{h}_c$ is defined as the concatenation of the graph embedding $\bar{\bm{h}}$, the embedding of the last node $\bm{h}_{\bm{\pi}_{t-1}}$, and the remaining vehicle capacity. For MOKP, $\bm{h}_c$ is defined as the concatenation of the graph embedding $\bar{\bm{h}}$ and the remaining knapsack capacity.

A masking mechanism is adopted in each decoding step to ensure the solution feasibility. For MOTSP, the visited nodes are masked. For MOCVRP (MOKP), besides the visited nodes, those with a demand (weight) larger than the remaining vehicle (knapsack) capacity are also masked.

\section{Instance augmentation}

In the inference process, an instance can be transformed into other variants sharing the same optimal solutions, so as to augment the performance. An instance of Bi-CVRP has 8 transformations, and an instance of $M$-objective TSP has $8^M$ transformations \cite{lin22} due to the full transformation permutation of $M$ groups of 2-dimensional coordinates, where each group has 8 transformations \cite{kwo20}, \{$(x,y),(1-x,y),(x,1-y),(1-x,1-y),(y,x),(1-y,x),(y,1-x),(1-y,1-x)$\}.

Our NHDE can effectively enhance the performance only using partial instance augmentation, which can reduce the solving time, since it can already achieve high diversity. Specifically, for $M$-objective TSP, NHDE adopts the full permutation of the first 4 transformations and last 4 transformations, respectively, thereby a total of $2 \times 4^M$ transformations.

\section{Training and fine-tuning of NHDE-M}

The training algorithm of NHDE-M in the meta-learning process, adapted from that of MDRL \cite{zha22}, is outlined in Algorithm \ref{train-NHDE-M}. Also, the three key adjustments based on MDRL are captured in Line 7, Line13, and Line 18. In the fine-tuning process, for the given $N$ weights $\bm{\lambda}$ as well as diversity factors $\bm{w}$, $N$ submodels are fine-tuned from the well-trained meta-model to solve the MOCO problem. The fine-tuning algorithm is presented in Algorithm \ref{fine-NHDE-M}.

\begin{algorithm}[!t]
	\caption{Training algorithm of NHDE-M}
	\label{train-NHDE-M}
	\begin{algorithmic}[1]
		\STATE \textbf{Input:} weight distribution $\Lambda$, diversity-factor distribution $\mathcal{W}$, instance distribution $\mathcal{S}$, initial meta-learning rate $\epsilon_0$, number of meta-iterations $T_m$, number of sampling steps per meta-iteration $N'$, number of sampled weights per sampling step $\tilde{N}$, number of update steps of the submodel $E$, batch size $B$, instance size $n$
		\STATE Initialize the meta-model $\bm{\theta}$
            \STATE $\epsilon \leftarrow \epsilon_0$
		\FOR {$t_m = 1$ to $T_m$}
            \STATE $s_{e,i} \sim \textbf{SampleInstance}(\mathcal{S}) \quad \forall e \in \{1,\cdots,E\} \quad \forall i \in \{1,\cdots,B\}$
            \STATE Initialize $\mathcal{F}_{e,i} \leftarrow \emptyset \quad \forall e,i$
            \FOR{$n' = 1$ to $N'$}
            \FOR{$\tilde{n} = 1$ to $\tilde{N}$}
            \STATE $\bm{\lambda} \sim \textbf{SampleWeight}(\Lambda)$
		\STATE $\bm{w} \sim \textbf{SampleDiversityFactor}(\mathcal{W})$
		\FOR {$e = 1$ to $E$}
            \STATE $\bm{\pi}_{i}^{j} \sim \textbf{SampleSolution}(P_{\bm{\theta}^{\tilde{n}}}(\cdot|s_{e,i},\tilde{\mathcal{F}}_{\bm{r},e,i})) \quad \forall i \in \{1,\cdots,B\} \quad \forall j \in \{1,\cdots,n\}$
            \STATE $R_i^{j} \leftarrow -w_1 g(\bm{\pi}^{j}_i|s_{e,i},\bm{\lambda})+ w_2 {\rm{HV}}_{\bm{r}}  (\tilde{\mathcal{F}}_{e,i}\cup\{\bm{\pi}^{j}_i\}) \quad \forall i,j$
		\STATE $b_i \leftarrow \frac{1}{n} \sum_{j=1}^{n} (-R_i^j) \quad \forall i$
		\STATE $\nabla \mathcal{J}(\bm{\theta}^{\tilde{n}}) \leftarrow \frac{1}{Bn} \sum_{i=1}^{B} \sum_{j=1}^{n} [ (-R_i^j - b_i) \nabla_{\bm{\theta}^{\tilde{n}}} \log P_{\bm{\theta}^{\tilde{n}}}(\bm{\pi}_{i}^{j}|s_i,\tilde{\mathcal{F}}_{\bm{r},i})]$
		\STATE $\bm{\theta}^{\tilde{n}} \leftarrow \textbf{Adam}(\bm{\theta}^{\tilde{n}},\nabla \mathcal{J}(\bm{\theta}^{\tilde{n}}))$
            \STATE $\mathcal{G}_i \leftarrow \{\bm{\pi}_{i}^{1},\dots,\bm{\pi}_{i}^{n}\} \quad \forall i$
            \STATE $\mathcal{F}_{e,i} \leftarrow {\rm{MPO}}(\tilde{\mathcal{F}}_{e,i} \cup \tilde{\mathcal{G}}_i) \quad \forall e,i$
		\ENDFOR
            \ENDFOR
		\STATE $\bm{\theta} \leftarrow \bm{\theta}+\epsilon(\frac{1}{\tilde{N}}\sum_{\tilde{n}=1}^{\tilde{N}} \bm{\theta}^{\tilde{n}}-\bm{\theta})$
		\STATE $\epsilon \leftarrow \epsilon-\epsilon_0/(T_m \times N')$
            \ENDFOR
            \ENDFOR 
		\STATE \textbf{Output:} The parameter of the meta-model $\bm{\theta}$
	\end{algorithmic}  
\end{algorithm}

\begin{algorithm}[!t]
	\caption{Fine-tuning algorithm of NHDE-M}
	\label{fine-NHDE-M}
	\begin{algorithmic}[1]
		\STATE \textbf{Input:} instance distribution $\mathcal{S}$, weights $\bm{\lambda}^1, \dots, \bm{\lambda}^N$, diversity factors $\bm{w}^1, \dots, \bm{w}^N$, number of fine-tuning steps of the submodel $E_f$, batch size $B$, instance size $n$, well-trained meta-model $\bm{\theta}$~
            \STATE $s_{e,i} \sim \textbf{SampleInstance}(\mathcal{S}) \quad \forall e \in \{1,\cdots,E_f\} \quad \forall i \in \{1,\cdots,B\}$
            \STATE Initialize $\mathcal{F}_{e,i} \leftarrow \emptyset \quad \forall e,i$
            \FOR{$\tilde{n} = 1$ to $N$}
            \STATE $\bm{\theta}^{\tilde{n}} \leftarrow \bm{\theta}$
		\FOR {$e = 1$ to $E$}
            \STATE $\bm{\pi}_{i}^{j} \sim \textbf{SampleSolution}(P_{\bm{\theta}^{\tilde{n}}}(\cdot|s_{e,i},\tilde{\mathcal{F}}_{\bm{r},e,i})) \quad \forall i \in \{1,\cdots,B\} \quad \forall j \in \{1,\cdots,n\}$
            \STATE $R_i^{j} \leftarrow -w_1 g(\bm{\pi}^{j}_i|s_{e,i},\bm{\lambda})+ w_2 {\rm{HV}}_{\bm{r}}  (\tilde{\mathcal{F}}_{e,i}\cup\{\bm{\pi}^{j}_i\}) \quad \forall i,j$
		\STATE $b_i \leftarrow \frac{1}{n} \sum_{j=1}^{n} (-R_i^j) \quad \forall i$
		\STATE $\nabla \mathcal{J}(\bm{\theta}^{\tilde{n}}) \leftarrow \frac{1}{Bn} \sum_{i=1}^{B} \sum_{j=1}^{n} [ (-R_i^j - b_i) \nabla_{\bm{\theta}^{\tilde{n}}} \log P_{\bm{\theta}^{\tilde{n}}}(\bm{\pi}_{i}^{j}|s_i,\tilde{\mathcal{F}}_{\bm{r},i})]$
		\STATE $\bm{\theta}^{\tilde{n}} \leftarrow \textbf{Adam}(\bm{\theta}^{\tilde{n}},\nabla \mathcal{J}(\bm{\theta}^{\tilde{n}}))$
            \STATE $\mathcal{G}_i \leftarrow \{\bm{\pi}_{i}^{1},\dots,\bm{\pi}_{i}^{n}\} \quad \forall i$
            \STATE $\mathcal{F}_{e,i} \leftarrow {\rm{MPO}}(\tilde{\mathcal{F}}_{e,i} \cup \tilde{\mathcal{G}}_i) \quad \forall e,i$
		\ENDFOR
            \ENDFOR 
		\STATE \textbf{Output:} The parameters of the fine-tuned submodels $\bm{\theta}^1, \dots, \bm{\theta}^N$
	\end{algorithmic}  
\end{algorithm}

\section{Hyperparameters of NHDE-M}

NHDE-M trains a meta-model with 150 meta-iterations and initial meta-learning rate $\epsilon_0=1$. We set $N'=20$, $\tilde{N}=M$, and $E=100$. We use batch size $B=64$ and the Adam \cite{kin15} optimizer with learning rate $10^{-4}$. During fine-tuning and inference, the number of fine-tuning steps $E_f$ is set to 50, and Adam with learning rate $10^{-4}$ is used. $N\!=\!40$ and $N\!=\!210$ uniformly distributed weights are generated and then shuffled for $M\!=\!2$ and $M\!=\!3$, respectively. $N$ diversity factors linearly change from (1,0) to (0,1). For the compared MDRL, the settings are the same as NHDE-M except 3000 meta-iterations are used, so that MDRL and NHDE-M execute the same number of gradient steps.

\section{Experimental results of NHDE-M}

\subsection{More results of NHDE-M}

NHDE-M usually spends relatively more inference time than MDRL with the same number of weights. Hence, we adjust the number of weights of MDRL, making its inference time close to or longer than NHDE-M for fair comparisons, as shown in Table \ref{tab:main-m-time}. Nonetheless, NHDE-M is still superior to MDRL in most cases. Without instance augmentation (aug.), NHDE-M is inferior to MDRL on Bi-CVRP50, Bi-CVRP100, and Tri-TSP100. Instance augmentation can effectively boost the performance of NHDE-M, where NHDE-M (aug.) is only inferior to MDRL (aug.) on Bi-CVRP100. Note that, for a given weight, MDRL needs to further fine-tune the meta-model with $E_f=50$ gradient steps to derive a submodel to the corresponding subproblem, which means that the increasing number of weights would cause considerable extra fine-tuning costs.

\begin{table}[!t]
  \centering
  \caption{Results of NHDE-M compared with MDRL with close or more total solving time on 200 random instances of MOCO problems.}
  \resizebox{0.99\textwidth}{!}{
  	\addtolength{\tabcolsep}{-4pt}
    \begin{tabular}{l|cccc|cccc|cccc}
    \toprule
          & \multicolumn{4}{c|}{Bi-TSP20} & \multicolumn{4}{c|}{Bi-TSP50} & \multicolumn{4}{c}{Bi-TSP100} \\
    Method & HV$\uparrow$    & $|$NDS$|$$\uparrow$ & Gap$\downarrow$   & Time  & HV$\uparrow$    & $|$NDS$|$$\uparrow$ & Gap$\downarrow$   & Time  & HV$\uparrow$    & $|$NDS$|$$\uparrow$ & Gap$\downarrow$   & Time \\
    \midrule
    MDRL (40 wt.) & 0.6264  & 20    & 0.49\% & 2s    & 0.6342  & 33    & 1.35\% & 3s    & 0.6940  & 36    & 1.55\% & 8s \\
    MDRL (600 wt.) & \underline{0.6287}  & 54    & \underline{0.13\%} & 29s   & 0.6380  & 133   & 0.76\% & 64s   & 0.7006  & 185   & 0.61\% & 2.1m \\
    NHDE-M (40 wt.) & \underline{0.6287}  & 58    & \underline{0.13\%} & 20s   & 0.6393  & 132   & 0.56\% & 57s   & 0.7008  & 195   & 0.58\% & 2.0m \\
    \midrule
    MDRL (40 wt. aug.) & 0.6267  & 18    & 0.44\% & 21s   & 0.6384  & 34    & 0.70\% & 1.5m  & 0.6995  & 38    & 0.77\% & 3.3m \\
    MDRL (100 wt. aug.) & 0.6271  & 23    & 0.38\% & 1.2m  & \underline{0.6408}  & 67    & \underline{0.33\%} & 3.6m  & \underline{0.7023}  & 82    & \underline{0.37\%} & 16m \\
    NHDE-M (40 wt. aug.) & \textbf{0.6295} & 81    & \textbf{0.00\%} & 1.5m  & \textbf{0.6429} & 273   & \textbf{0.00\%} & 2.6m  & \textbf{0.7049} & 339   & \textbf{0.00\%} & 5.5m \\
    \midrule
          & \multicolumn{4}{c|}{Bi-CVRP20} & \multicolumn{4}{c|}{Bi-CVRP50} & \multicolumn{4}{c}{Bi-CVRP100} \\
    Method & HV$\uparrow$    & $|$NDS$|$$\uparrow$ & Gap$\downarrow$   & Time  & HV$\uparrow$    & $|$NDS$|$$\uparrow$ & Gap$\downarrow$   & Time  & HV$\uparrow$    & $|$NDS$|$$\uparrow$ & Gap$\downarrow$   & Time \\
    \midrule
    MDRL (40 wt.) & 0.4284  & 9     & 0.49\% & 3s    & 0.4057  & 5     & 1.12\% & 5s    & 0.4015  & 0     & 1.45\% & 10s \\
    MDRL (300 wt.) & 0.4296  & 17    & 0.21\% & 23s   & \underline{0.4089}  & 21    & \underline{0.34\%} & 49s   & \underline{0.4078}  & 21    & \underline{-0.10\%} & 1.5m \\
    NHDE-M (40 wt.) & 0.4296  & 16    & 0.21\% & 23s   & 0.4086  & 20    & 0.41\% & 47s   & 0.4053  & 18    & 0.52\% & 1.4m \\
    \midrule
    MDRL (40 wt. aug.) & 0.4293  & 9     & 0.28\% & 5s    & 0.4073  & 11    & 0.73\% & 16s   & 0.4040  & 11    & 0.83\% & 1.0m \\
    MDRL (300 wt. aug.) & \underline{0.4302}  & 16    & \underline{0.07\%} & 1.0m  & \textbf{0.4103} & 24    & \textbf{0.00\%} & 2.1m  & \textbf{0.4086} & 24    & \textbf{-0.29\%} & 7.7m \\
    NHDE-M (40 wt. aug.) & \textbf{0.4305} & 24    & \textbf{0.00\%} & 1.2m  & \textbf{0.4103} & 29    & \textbf{0.00\%} & 1.6m  & 0.4074  & 26    & 0.00\% & 2.7m \\
    \midrule
          & \multicolumn{4}{c|}{Bi-KP50}  & \multicolumn{4}{c|}{Bi-KP100} & \multicolumn{4}{c}{Bi-KP200} \\
    Method & HV$\uparrow$    & $|$NDS$|$$\uparrow$ & Gap$\downarrow$   & Time  & HV$\uparrow$    & $|$NDS$|$$\uparrow$ & Gap$\downarrow$   & Time  & HV$\uparrow$    & $|$NDS$|$$\uparrow$ & Gap$\downarrow$   & Time \\
    \midrule
    MDRL (40 wt.) & 0.3559  & 17    & 0.20\% & 4s    & 0.4528  & 25    & 0.31\% & 8s    & 0.3594  & 31    & 0.42\% & 24s \\
    MDRL (300 wt.) & \underline{0.3563}  & 29    & \underline{0.08\%} & 30s   & \underline{0.4536}  & 58    & \underline{0.13\%} & 1.0m  & \underline{0.3606}  & 95    & \underline{0.08\%} & 3.1m \\
    NHDE-M (40 wt.) & \textbf{0.3566} & 41    & \textbf{0.00\%} & 31s   & \textbf{0.4542} & 93    & \textbf{0.00\%} & 1.0m  & \textbf{0.3609} & 160   & \textbf{0.00\%} & 2.8m \\
    \midrule
          & \multicolumn{4}{c|}{Tri-TSP20} & \multicolumn{4}{c|}{Tri-TSP50} & \multicolumn{4}{c}{Tri-TSP100} \\
    Method & HV$\uparrow$    & $|$NDS$|$$\uparrow$ & Gap$\downarrow$   & Time  & HV$\uparrow$    & $|$NDS$|$$\uparrow$ & Gap$\downarrow$   & Time  & HV$\uparrow$    & $|$NDS$|$$\uparrow$ & Gap$\downarrow$   & Time \\
    \midrule
    MDRL (210 wt.) & 0.4723  & 126   & 0.90\% & 14s   & 0.4388  & 199   & 4.46\% & 20s   & 0.4956  & 207   & 3.17\% & 40s \\
    MDRL (3003 wt.) & 0.4761  & 479   & 0.10\% & 2.6m  & \underline{0.4512}  & 1927  & \underline{1.76\%} & 5.1m  & \underline{0.5104}  & 2400  & \underline{0.27\%} & 10m \\
    NHDE-M (210 wt.) & \underline{0.4763}  & 783   & \underline{0.06\%} & 1.4m  & \underline{0.4512}  & 2636  & \underline{1.76\%} & 4.7m  & 0.4997  & 4056  & 2.36\% & 11m \\
    \midrule
    MDRL (210 wt. aug.) & 0.4727  & 107   & 0.82\% & 14m   & 0.4473  & 202   & 2.61\% & 53m   & 0.5056  & 209   & 1.21\% & 4.3h \\
    MDRL (153 wt. aug.) & 0.4721  & 92    & 0.94\% & 11m   & 0.4448  & 150   & 3.16\% & 39m   & 0.5022  & 153   & 1.88\% & 3.2h \\
    NHDE-M (210 wt. aug.) & \textbf{0.4766} & 748   & \textbf{0.00\%} & 13m   & \textbf{0.4593} & 10850  & \textbf{0.00\%} & 30m   & \textbf{0.5118} & 13216  & \textbf{0.00\%} & 1.5h \\
    \bottomrule
    \end{tabular}%
    }
  \label{tab:main-m-time}%
\end{table}%

\subsection{Generalization study of NHDE-M}

\begin{figure}[!t]
	\centering
	\subfigure[]{
		\centering
		\includegraphics[width=0.31\textwidth]{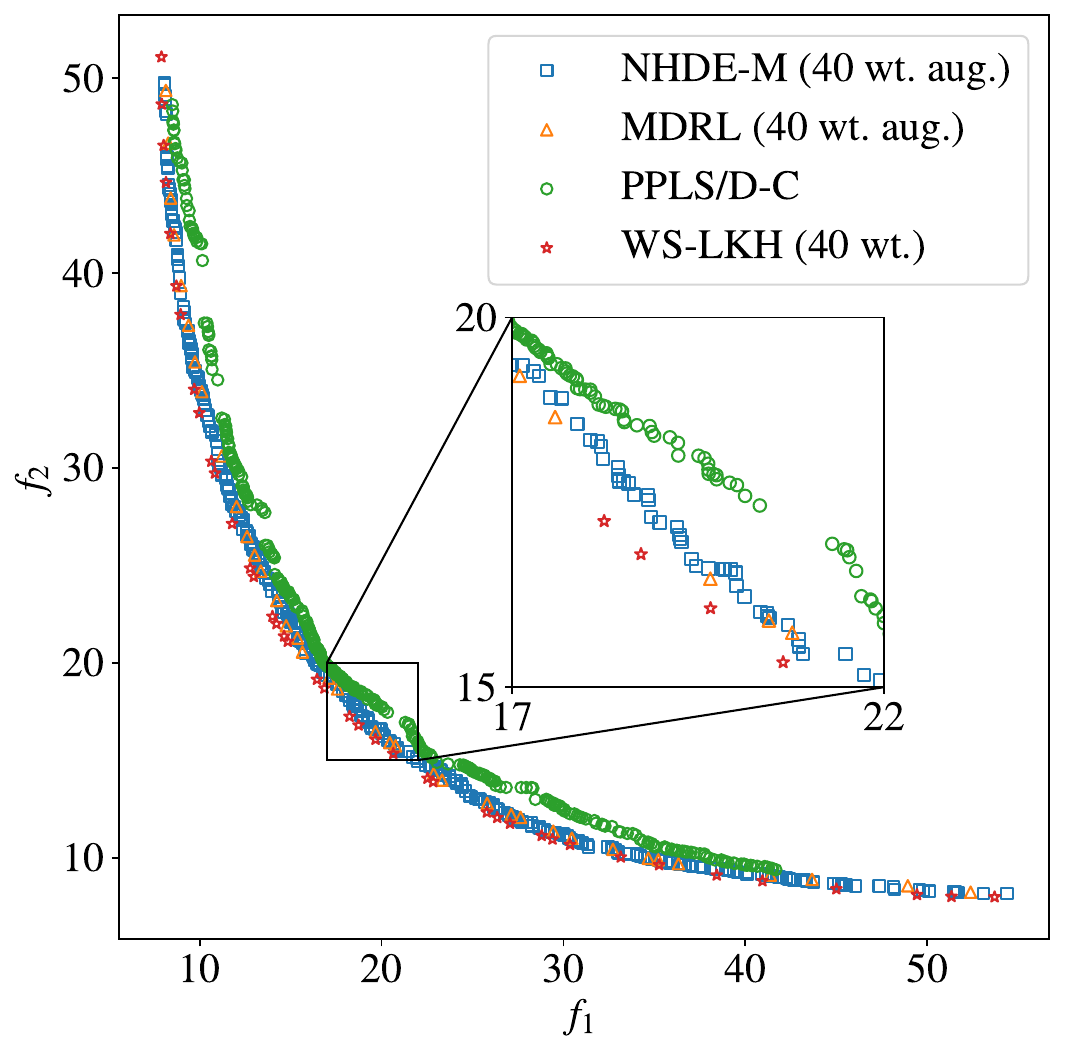}
		\label{fig:kro100}
	}
	\subfigure[]{
		\centering
		\includegraphics[width=0.31\textwidth]{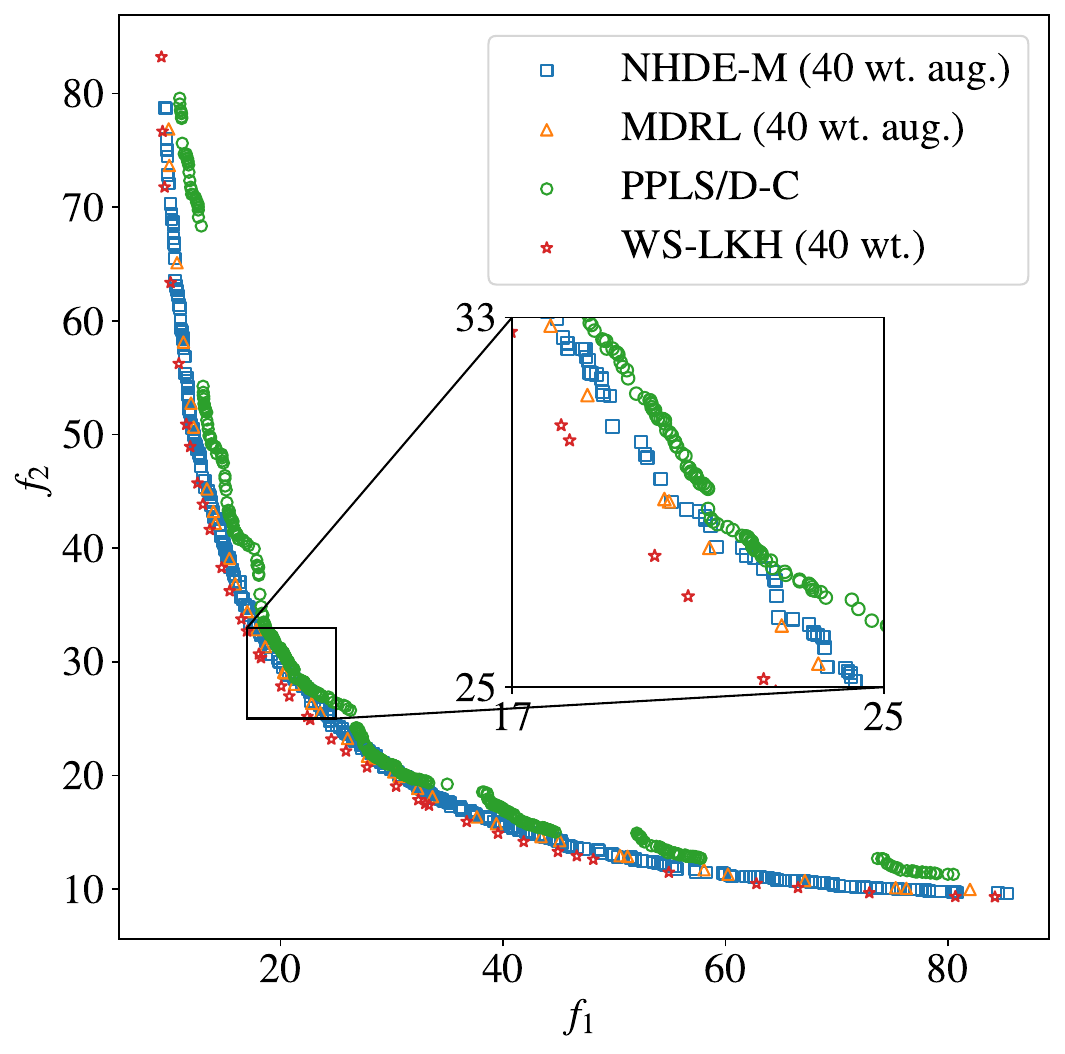}
		\label{fig:kro150}
	}
	\subfigure[]{
		\centering
		\includegraphics[width=0.31\textwidth]{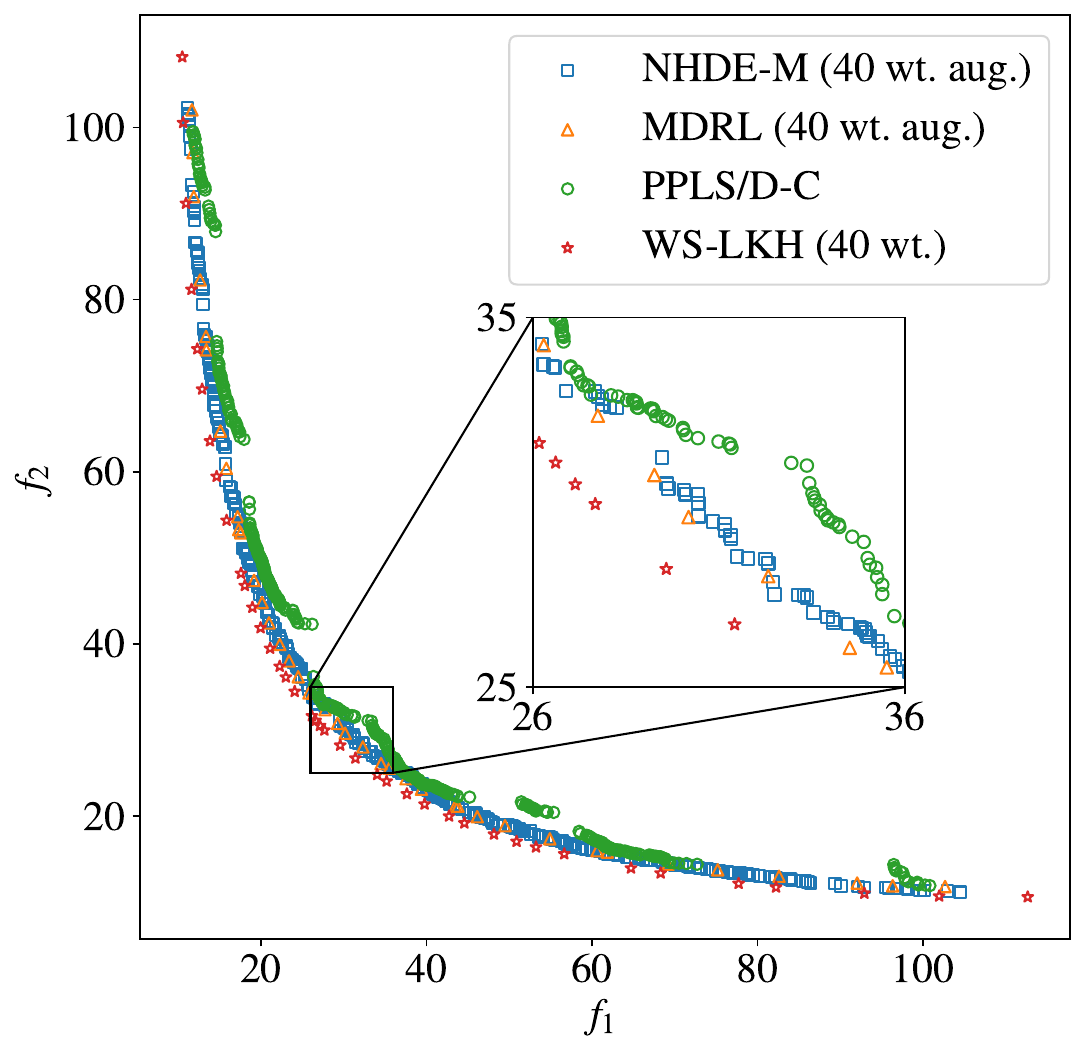}
		\label{fig:kro200}
	}
	\caption{Pareto fronts of NHDE-M and compared methods on benchmark instances. (a) KroAB100. (b) KroAB150. (c) KroAB200.}
	\label{fig:kro-m}
\end{figure}

We assess the zero-shot generalization capability of NHDE-M, which is trained and fine-tuned on Bi-TSP100, and tested on 200 random larger-scale Bi-TSP instances, i.e., Bi-TSP150/200, and three commonly used benchmark instances, i.e., KroAB100/150/200. The results are gathered in Tables \ref{tab:gen-m} and \ref{tab:kro-m}. The Pareto fronts of the benchmark instances obtained by various methods are also visualized in Figure \ref{fig:kro-m}. As can be clearly observed, NHDE-M exhibits superior generalization capability to the state-of-the-art MOEA and other neural methods with regard to the convergence and diversity.

\begin{table}[!t]
  \centering
  \caption{Results of NHDE-M on 200 random instances of larger-scale problems.}
    \resizebox{0.9\textwidth}{!}{
    \begin{tabular}{l|cccc|cccc}
    \toprule
          & \multicolumn{4}{c|}{Bi-TSP150} & \multicolumn{4}{c}{Bi-TSP200} \\
    Method & HV$\uparrow$    & $|$NDS$|$$\uparrow$ & Gap$\downarrow$   & Time  & HV$\uparrow$    & $|$NDS$|$$\uparrow$ & Gap$\downarrow$   & Time \\
    \midrule
    WS-LKH (40 wt.) & \textbf{0.7075} & 39    & \textbf{-0.90\%} & 5.3h  & \textbf{0.7435} & 40    & \textbf{-1.38\%} & 8.5h \\
    PPLS/D-C (200 iter.) & 0.6784  & 473   & 3.25\% & 21h   & 0.7106  & 512   & 3.11\% & 32h \\
    DRL-MOA (101 models) & 0.6901  & 73    & 1.58\% & 45s   & 0.7219  & 75    & 1.57\% & 87s \\
    \midrule
    MDRL (40 wt.) & 0.6894  & 37    & 1.68\% & 23s   & 0.7227  & 38    & 1.46\% & 42s \\
    MDRL (400 wt.) & 0.6958  & 176   & 0.77\% & 3.7m  & 0.7284  & 190   & 0.68\% & 6.9m \\
    NHDE-M (40 wt.) & 0.6970  & 239   & 0.60\% & 3.0m  & 0.7297  & 268   & 0.50\% & 4.1m \\
    \midrule
    MDRL (40 wt. aug.) & 0.6948  & 39    & 0.91\% & 21m   & 0.7275  & 39    & 0.80\% & 41m \\
    NHDE-M (40 wt. aug.) & \underline{0.7012}  & 373   & \underline{0.00\%} & 14m   & \underline{0.7334}  & 395   & \underline{0.00\%} & 25m \\
    \bottomrule
    \end{tabular}%
    }
  \label{tab:gen-m}%
\end{table}%

\begin{table}[!t]
  \centering
  \caption{Results of NHDE-M on benchmark instances.}
    \resizebox{0.99\textwidth}{!}{
    \addtolength{\tabcolsep}{-4pt}
    \begin{tabular}{l|cccc|cccc|cccc}
    \toprule
          & \multicolumn{4}{c|}{KroAB100} & \multicolumn{4}{c|}{KroAB150} & \multicolumn{4}{c}{KroAB200} \\
    Method & HV$\uparrow$    & $|$NDS$|$$\uparrow$ & Gap$\downarrow$   & Time  & HV$\uparrow$    & $|$NDS$|$$\uparrow$ & Gap$\downarrow$   & Time  & HV$\uparrow$    & $|$NDS$|$$\uparrow$ & Gap$\downarrow$   & Time \\
    \midrule
    WS-LKH (40 wt.) & \textbf{0.7007} & 40    & \textbf{-0.42\%} & 53s   & \textbf{0.6989} & 39    & \textbf{-0.92\%} & 1.9m  & \textbf{0.7404} & 40    & \textbf{-1.48\%} & 2.2m \\
    PPLS/D-C (200 iter.) & 0.6785  & 388   & 2.77\% & 31m   & 0.6659  & 441   & 3.84\% & 1.1h  & 0.7100  & 491   & 2.69\% & 3.1h \\
    DRL-MOA (101 models) & 0.6903  & 67    & 1.07\% & 10s   & 0.6794  & 72    & 1.89\% & 18s   & 0.7185  & 73    & 1.52\% & 23s \\
    \midrule
    MDRL (40 wt.) & 0.6869  & 37    & 1.56\% & 5s    & 0.6810  & 36    & 1.66\% & 9s    & 0.7184  & 39    & 1.54\% & 12s \\
    NHDE-M (40 wt.) & 0.6940  & 183   & 0.54\% & 7s    & 0.6879  & 232   & 0.66\% & 12s   & 0.7253  & 275   & 0.59\% & 16s \\
    \midrule
    MDRL (40 wt. aug.) & 0.6928  & 37    & 0.72\% & 7s    & 0.6857  & 38    & 0.98\% & 10s   & 0.7234  & 40    & 0.85\% & 15s \\
    NHDE-M (40 wt. aug.) & \underline{0.6978}  & 341   & \underline{0.00\%} & 10s   & \underline{0.6925}  & 370   & \underline{0.00\%} & 12s   & \underline{0.7296}  & 393   & \underline{0.00\%} & 16s \\
    \bottomrule
    \end{tabular}%
    }
  \label{tab:kro-m}%
\end{table}%

\section{Detailed results of NHDE-P on benchmark instances}

Table \ref{tab:kro} records the detailed results of NHDE-P and other baselines on benchmark instances, which demonstrate the superiority of NHDE-P.

\begin{table}[!t]
  \centering
  \caption{Results of NHDE-P on benchmark instances.}
    \resizebox{0.99\textwidth}{!}{
    \addtolength{\tabcolsep}{-4pt}
    \begin{tabular}{l|cccc|cccc|cccc}
    \toprule
          & \multicolumn{4}{c|}{KroAB100} & \multicolumn{4}{c|}{KroAB150} & \multicolumn{4}{c}{KroAB200} \\
    Method & HV$\uparrow$    & $|$NDS$|$$\uparrow$ & Gap$\downarrow$   & Time  & HV$\uparrow$    & $|$NDS$|$$\uparrow$ & Gap$\downarrow$   & Time  & HV$\uparrow$    & $|$NDS$|$$\uparrow$ & Gap$\downarrow$   & Time \\
    \midrule
    WS-LKH (40 wt.) & \textbf{0.7007} & 40    & \textbf{-0.47\%} & 53s   & \textbf{0.6989} & 39    & \textbf{-0.92\%} & 1.9m  & \textbf{0.7404} & 40    & \textbf{-1.58\%} & 2.2m \\
    PPLS/D-C (200 iter.) & 0.6785  & 388   & 2.71\% & 31m   & 0.6659  & 441   & 3.84\% & 1.1h  & 0.7100  & 491   & 2.59\% & 3.1h \\
    DRL-MOA (101 models) & 0.6903  & 67    & 1.02\% & 10s   & 0.6794  & 72    & 1.89\% & 18s   & 0.7185  & 73    & 1.43\% & 23s \\
    \midrule
    PMOCO (40 wt.) & 0.6862  & 36    & 1.61\% & 7s    & 0.6802  & 37    & 1.78\% & 9s    & 0.7174  & 38    & 1.58\% & 12s \\
    NHDE-P (40 wt.) & 0.6926  & 179   & 0.69\% & 8s    & 0.6873  & 225   & 0.75\% & 12s   & 0.7247  & 246   & 0.58\% & 16s \\
    \midrule
    PMOCO (40 wt. aug.) & 0.6916  & 38    & 0.83\% & 8s    & 0.6861  & 37    & 0.92\% & 11s   & 0.7223  & 39    & 0.91\% & 16s \\
    NHDE-P (40 wt. aug.) & \underline{0.6974}  & 317   & \underline{0.00\%} & 11s   & \underline{0.6925}  & 365   & \underline{0.00\%} & 13s   & \underline{0.7289}  & 377   & \underline{0.00\%} & 17s \\
    \bottomrule
    \end{tabular}%
    }
  \label{tab:kro}%
\end{table}%

\section{Details of compared methods in ablation study}

In Figure \ref{fig:abla}, we compare NHDE-P with decomposition-based DRL without indicator (NHDE w/o I) and indicator-based DRL without decomposition (NHDE w/o D) to study the effect of the indicator-enhanced DRL. Concretely, NHDE w/o I removes the HV indicator in the reward and the Pareto front graph in the inputs. NHDE w/o D dispenses with weights, removes the scalar objective in the reward, and only adopts the HV indicator to guide the model. In each subproblem without a weight, a new solution (or multiple sampled solutions) is produced to maximize the HV indicator under the current Pareto front. The surrogate landscape cannot be defined due to the disuse of weights, so the current whole Pareto front is taken as the input to the model.

Since HV can comprehensively measure convergence and diversity, indicator-based NHDE w/o D should find a Pareto set with good overall performance in intuition. However, it is even inferior to decomposition-based NHDE w/o I in practice. This fact reveals that it is difficult for the deep model to learn to construct solutions to directly optimize HV due to the high complexity of HV.

With respect to NHDE w/o MPO, which is used to evaluate the impact of our MPO strategy, it still samples multiple solutions for each subproblem, but only the solution with the maximum reward is preserved according to the view of single-objective optimization.

\section{Runtime analysis}

We provide the NHDE and PMOCO results on Bi-TSP50 with similar runtime, by changing the number of weights used in both methods in Table \ref{tab:timeana}. When using a few weights (short runtime), PMOCO is slightly better than NHDE with close runtime (since the increasing number of weights can rapidly raise the performance of PMOCO). However, when using more weights ($N\geq300$ for PMOCO), NHDE is consistently better than PMOCO with a close runtime. This further verifies that increasing weights may not effectively produce more Pareto solutions for existing neural solvers, while our NHDE can boost the limitation of such decomposition-based neural solvers, especially in diversity. The superiority is more significant for larger and more complex MOCVRP instances.

\begin{table}[!t]
  \centering
  \caption{Results with change of the number of weights.}
    \begin{tabular}{cccccccc}
    \toprule
    \multicolumn{2}{c}{Number of wt.} & \multicolumn{2}{c}{Time} & \multicolumn{2}{c}{HV} & \multicolumn{2}{c}{$|$NDS$|$} \\
    NHDE-P & PMOCO & NHDE-P & PMOCO & NHDE-P & PMOCO & NHDE-P & PMOCO \\
    \midrule
    10    & 150   & 14s   & 14s   & 0.6334 & 0.6355 & 76    & 56 \\
    20    & 300   & 26s   & 26s   & 0.6373 & 0.6359 & 103   & 63 \\
    40    & 600   & 53s   & 53s   & 0.6388 & 0.6361 & 127   & 68 \\
    80    & 1200  & 1.3m  & 1.4m  & 0.6395 & 0.6362 & 146   & 70 \\
    \bottomrule
    \end{tabular}%
  \label{tab:timeana}%
\end{table}%

Moreover, we present the runtime proportion of each module in NHDE-P, including the original PMOCO, indicator-enhanced DRL, and MPO. The experiment is conducted on Bi-TSP50, as shown in Table \ref{tab:timeprop}. Please note that the runtime of the indicator-enhanced DRL during inference is mainly spent by the heterogeneous graph attention (HGA). The indicator-enhanced inference costs 5\% runtime, with the complexity of the attention mechanism in PMOCO being $O(n^2)$ and the additional complexity in HGA being $O(nK)$. MPO costs 82\% runtime, since the update of Pareto front needs more computation, i.e., $O((K+J)J)$.

\begin{table}[!t]
  \centering
  \caption{Results for the runtime proportion of each module.}
    \begin{tabular}{lc}
    \toprule
    Module of NHDE-P & Runtime Proportion \\
    \midrule
    PMOCO & 13\% \\
    Indicator-enhanced inference & 5\% \\
    MPO & 82\% \\
    \bottomrule
    \end{tabular}%
  \label{tab:timeprop}%
\end{table}%

We observe that the promising performance of NHDE comes more from MPO (see NHDE w/o MPO with HV 0.6335 in Figure \ref{fig:abla}) than the indicator-enhanced inference (see NHDE w/o I with HV 0.6361 in Figure \ref{fig:abla}). As presented in Table \ref{tab:timeprop}, the indicator-enhanced inference only costs a very small part of runtime, while MPO costs most of the runtime. Thus, their corresponding contributions to performance are reasonable, considering their compurational efforts.

We also compare NHDE-P with NHDE w/o I (equivalent to PMOCO with MPO) when using much more weights (i.e., 600) on Bi-TSP50. As shown in Table \ref{tab:600}, our NHDE outperforms PMOCO with MPO, where PMOCO spends similar runtime to NHDE with the extra MPO module.

\begin{table}[!t]
  \centering
  \caption{Comparison on NHDE-P with NHDE w/o I when using much more weights.}
    \begin{tabular}{lccc}
    \toprule
    Method & HV & $|$NDS$|$ & Time \\
    \midrule
    NHDE-P (600 wt.) & 0.6405 & 177 & 6.3m \\
    NHDE w/o I (600 wt.) & 0.6372 & 141 & 5.9m \\
    \bottomrule
    \end{tabular}%
  \label{tab:600}%
\end{table}%

\section{Weight assignment}

Recall that we use uniformly distributed weights during inference, which is a mainstream method for weight assignment when no information about the Pareto front is known in advance. For decomposition-based methods, weight assignment methods may affect the solution distribution. However, our NHDE can better alleviate this issue compared with pure decomposition-based neural heuristics for two reasons: (1) NHDE can generate more diverse solutions as verified by our experiments. (2) NHDE can also flexibly handle arbitrary weights during inference, enabling it to integrate seamlessly with proper weight assignment methods. When knowing the approximate scales of different objectives beforehand, we can first normalize them into [0,1] to derive a more uniform Pareto front. Otherwise, we can assign biased and non-uniform weights during inference to obtain more uniformly distributed solutions.

\begin{figure}[!t]
	\centering
	\subfigure[]{
		\centering
		\includegraphics[width=0.31\textwidth]{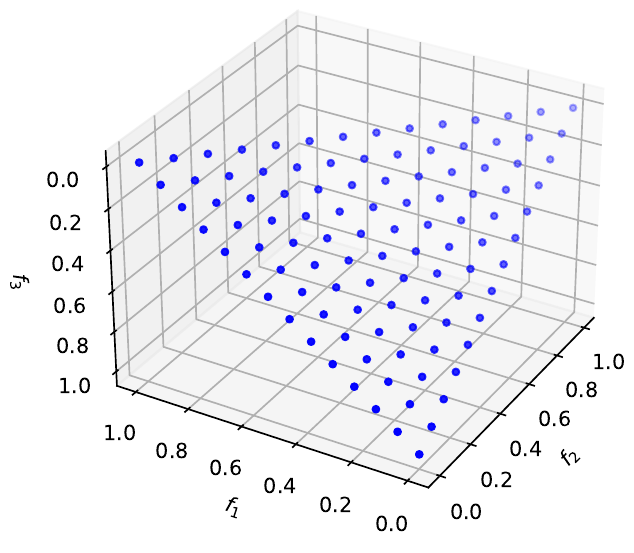}
		\label{fig:wtass-a}
	}
	\subfigure[]{
		\centering
		\includegraphics[width=0.31\textwidth]{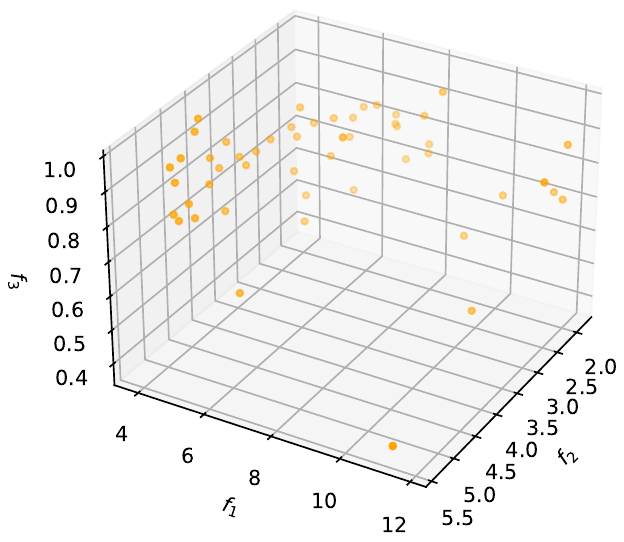}
		\label{fig:wtass-b}
	}
	\subfigure[]{
		\centering
		\includegraphics[width=0.31\textwidth]{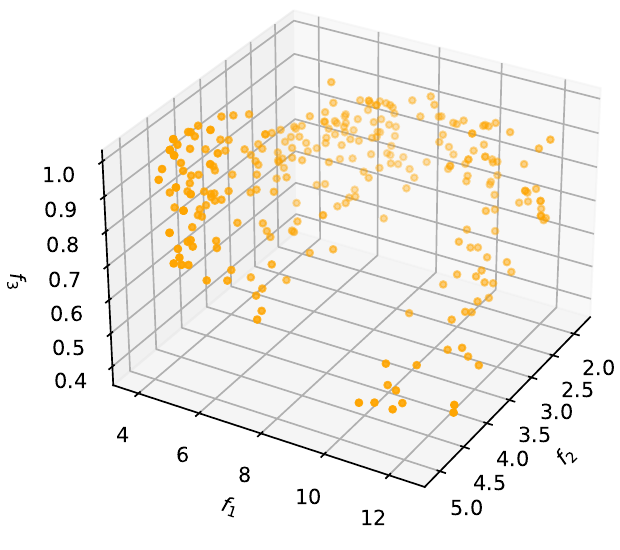}
		\label{fig:wtass-c}
	}
	\subfigure[]{
		\centering
		\includegraphics[width=0.31\textwidth]{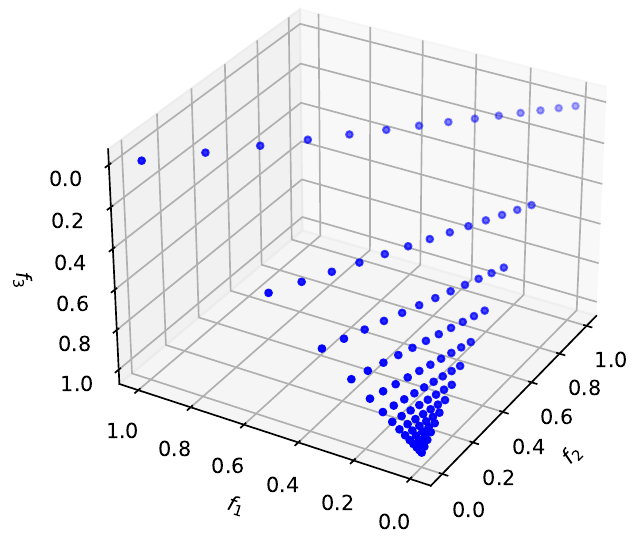}
		\label{fig:wtass-d}
	}
	\subfigure[]{
		\centering
		\includegraphics[width=0.31\textwidth]{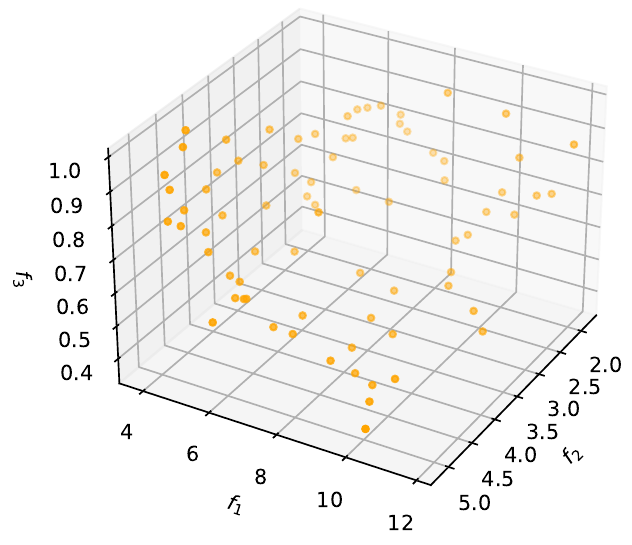}
		\label{fig:wtass-e}
	}
	\subfigure[]{
		\centering
		\includegraphics[width=0.31\textwidth]{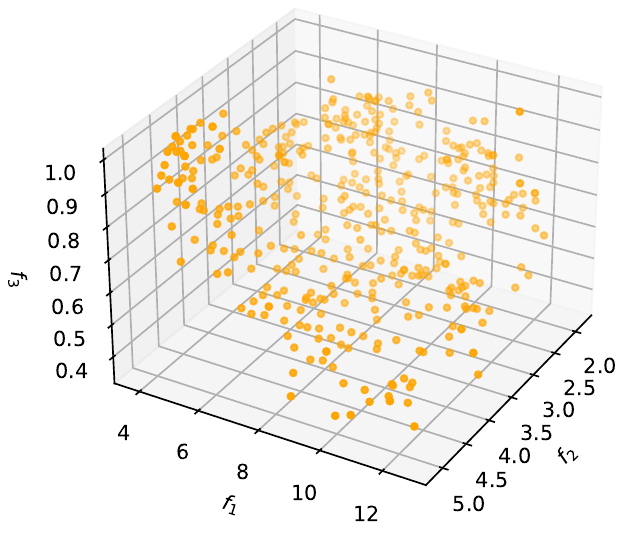}
		\label{fig:wtass-f}
	}
	\caption{Solutions generated by using 105 uniform/non-uniform distributed weights on an instance of Tri-TSP20 with asymmetric Pareto front. (a) Uniform weights. (b) PMOCO with uniform weights. (c) NHDE-P with uniform weights. (d) Non-uniform weights. (e) PMOCO with non-uniform weights. (f) NHDE-P with non-uniform weights.}
	\label{fig:wtass}
\end{figure}

We present the results on Tri-TSP with asymmetric Pareto fronts, as shown in Figure \ref{fig:wtass}. For Tri-TSP instances, the coordinates for the three objectives are randomly sampled from $[0,1]^2$, $[0,0.5]^2$, $[0,0.1]^2$, respectively. The results show that non-uniform weights, which are obtained by multiplying uniform weights by (1,2,10) element-wise and then normalizing them back to $[0,1]^3$, can produce a relatively more uniform Pareto front. Besides, compared with PMOCO (see Figure \ref{fig:wtass-b}), NHDE-P (see Figure \ref{fig:wtass-c}) can enhance diversity, thereby alleviating the non-uniform distribution of solutions.

\section{Hyperparameter study}

We further study the effects of $N'$ (the number of weights used in training), $K$ (the limited size of the surrogate landscape of the Pareto front), and $J$ (the limited number of \emph{points} from new solutions for updating the Pareto front).

\begin{figure}[!t]
	\centering
	\subfigure[]{
		\centering
		\includegraphics[width=0.31\textwidth]{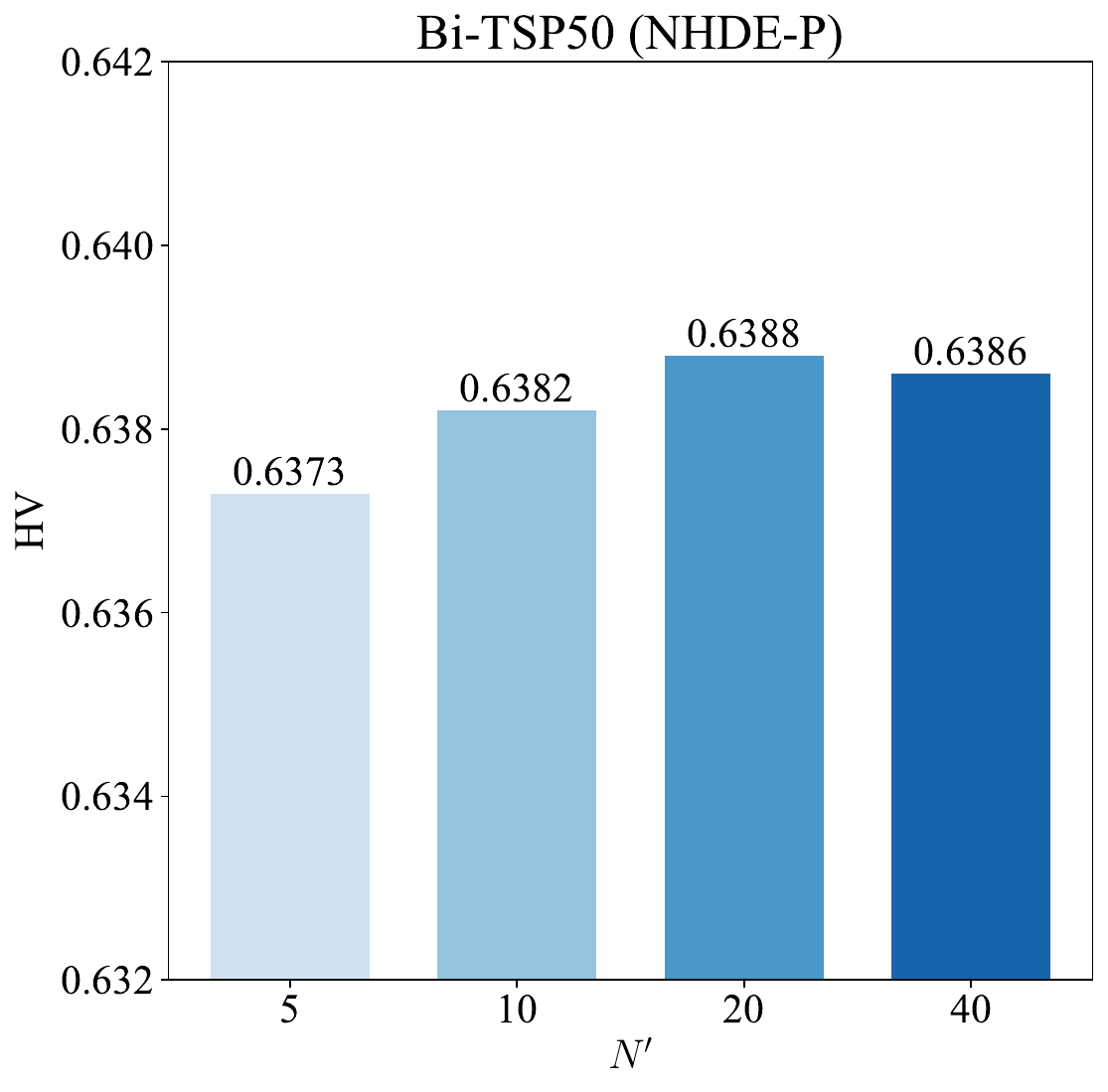}
		\label{fig:abla_npi}
	}
	\subfigure[]{
		\centering
		\includegraphics[width=0.31\textwidth]{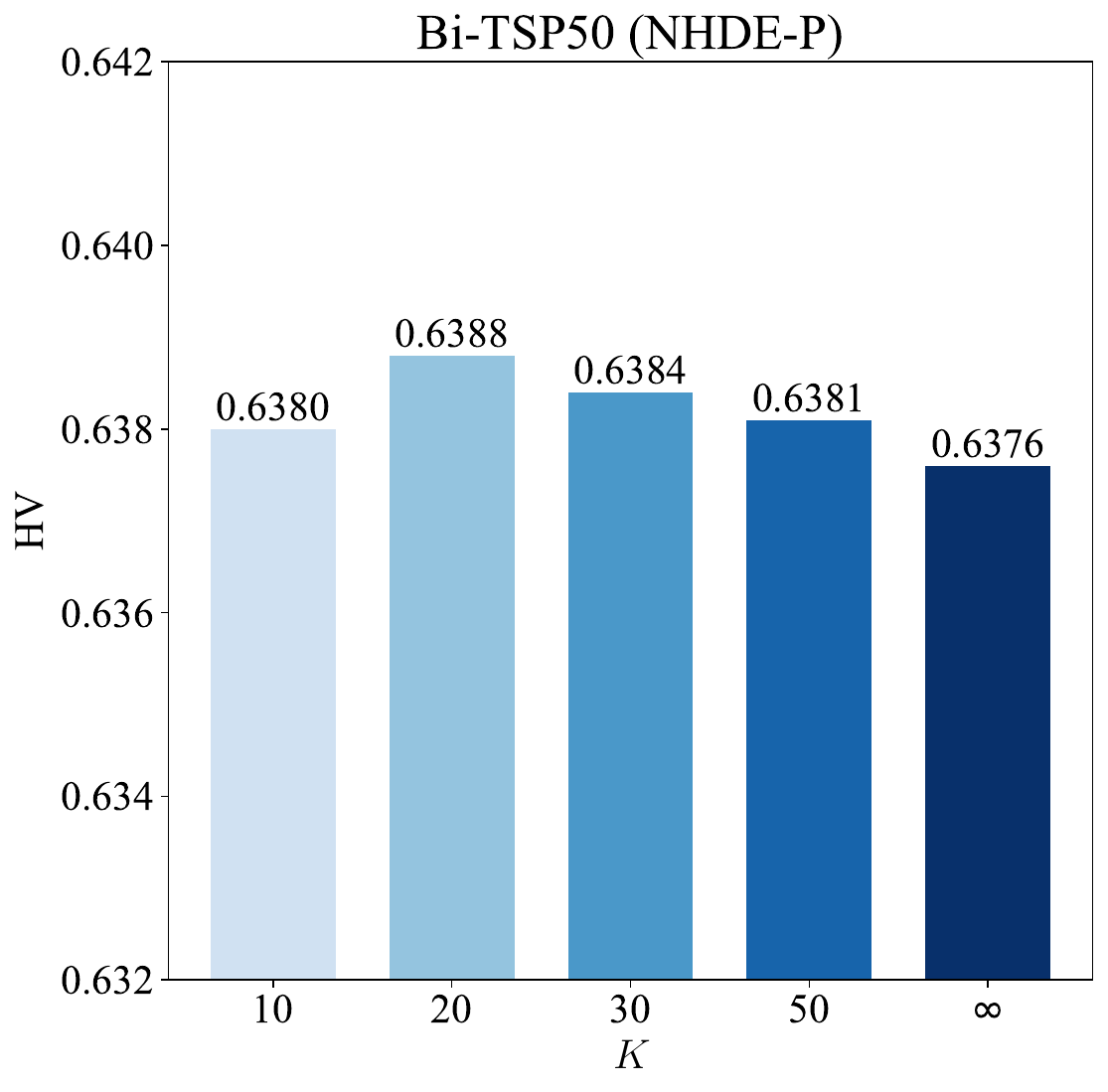}
		\label{fig:abla_k}
	}
	\subfigure[]{
		\centering
		\includegraphics[width=0.31\textwidth]{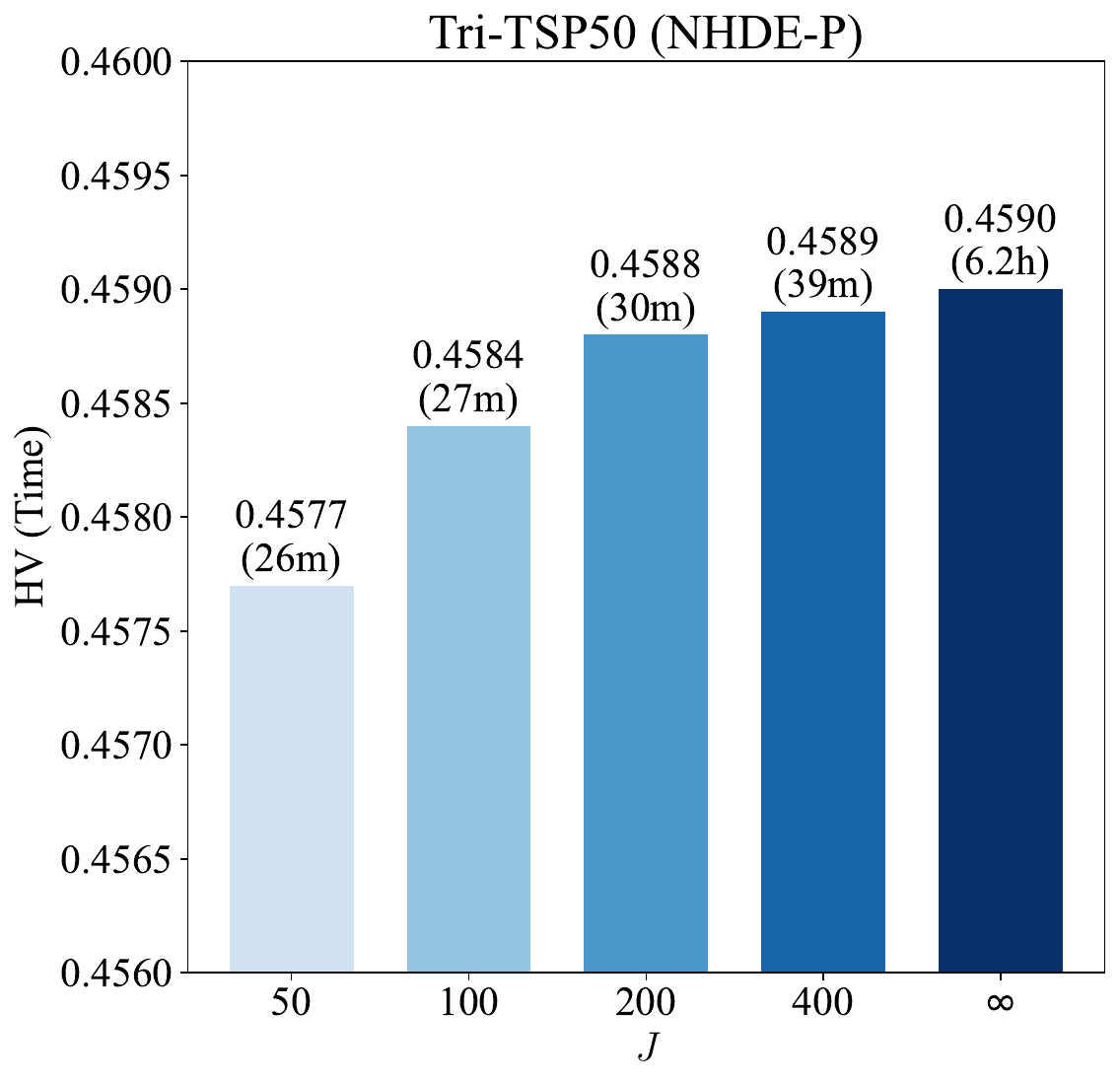}
		\label{fig:abla_j}
	}
	\caption{Hyperparameter study. (a) Effect of the number of weights used in training. (b) Effect of the limited size of the surrogate landscape of the Pareto front. (c) Effect of the limited number of \emph{points} from new solutions for updating the Pareto front.}
	\label{fig:hyperpara}
\end{figure}

We present the results of various $N'$ on Bi-TSP50 in the table below. As shown in Figure \ref{fig:abla_npi}, $N'=5$ and $N'=10$ cause inferior performance, while proper $N'$ ($20\leq N'\leq 40$) results in desirable performance. Intuitively, when limiting the same total gradient steps in training, a larger $N'$ means fewer instances used for model training. In this sense, too large $N'$, i.e., with insufficient instances, could lead to the inferior performance for solving unseen instances. On the other hand, too small $N'$ could prevent the model from learning favorable weight representations and thus deteriorate the final performance. Hence we choose $N'=20$ in this paper.

Figure \ref{fig:abla_k} displays the results of various values of $K$, where $K=20$ is a desirable setting. When $K$ is too small, some key information of the Pareto front would be lost, thereby degrading the performance. When $K$ is too large, the deep model cannot cope with numerous points of the Pareto front, also leading to deterioration of the performance.

We provide the HV and runtime of NHDE-P on Tri-TSP50 with the changed $J$ in Figure \ref{fig:abla_j}. We observe that by limiting $J$, the massive time of the update can be curtailed with only a little sacrifice of performance. Since $J=200$ is a good trade-off between HV and runtime, we use it in this paper.

\section{Additional analysis}

\begin{figure}[!t]
	\centering
	\subfigure[]{
		\centering
		\includegraphics[width=0.31\textwidth]{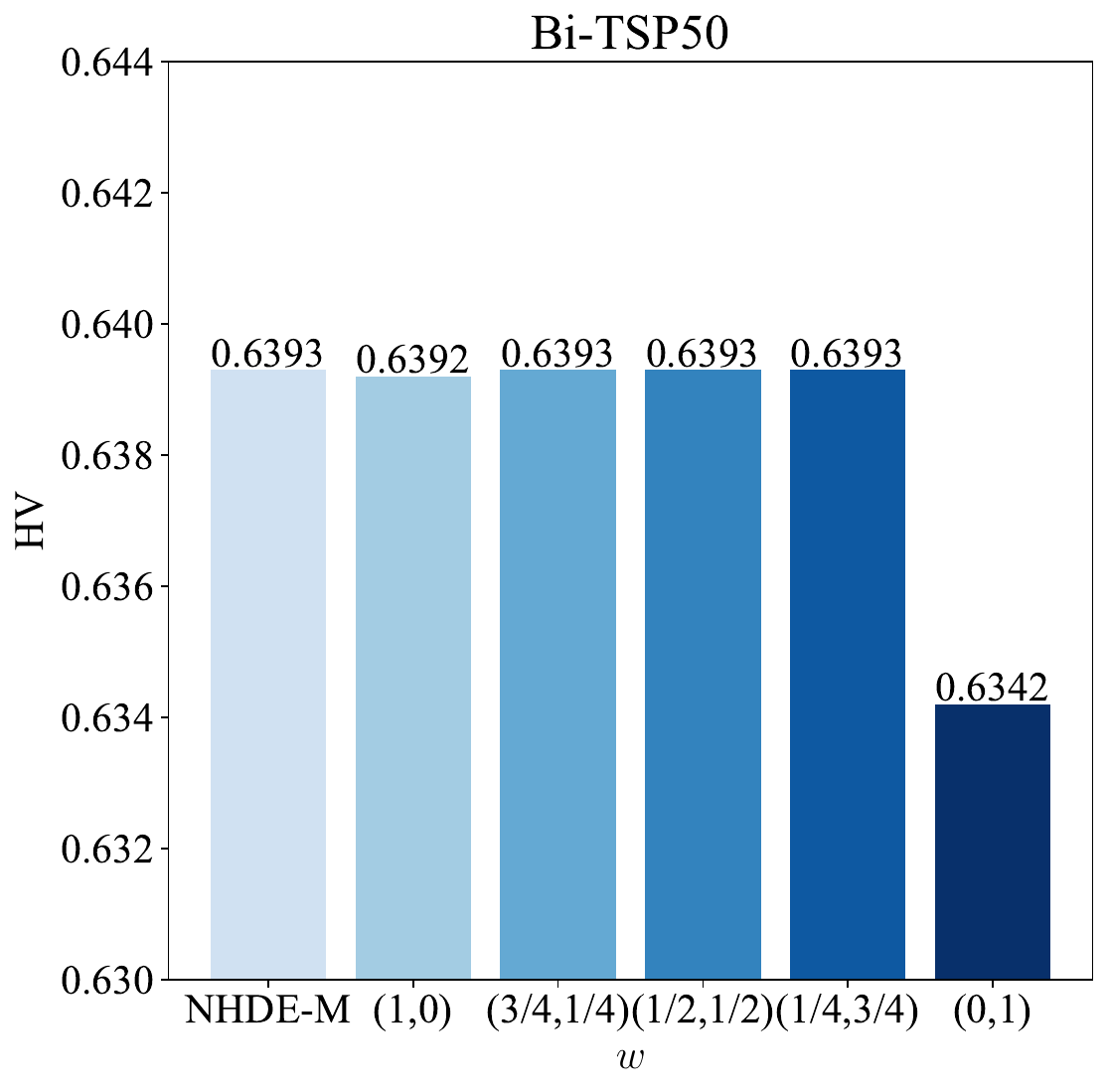}
		\label{fig:abla_w}
	}
	\subfigure[]{
		\centering
		\includegraphics[width=0.33\textwidth]{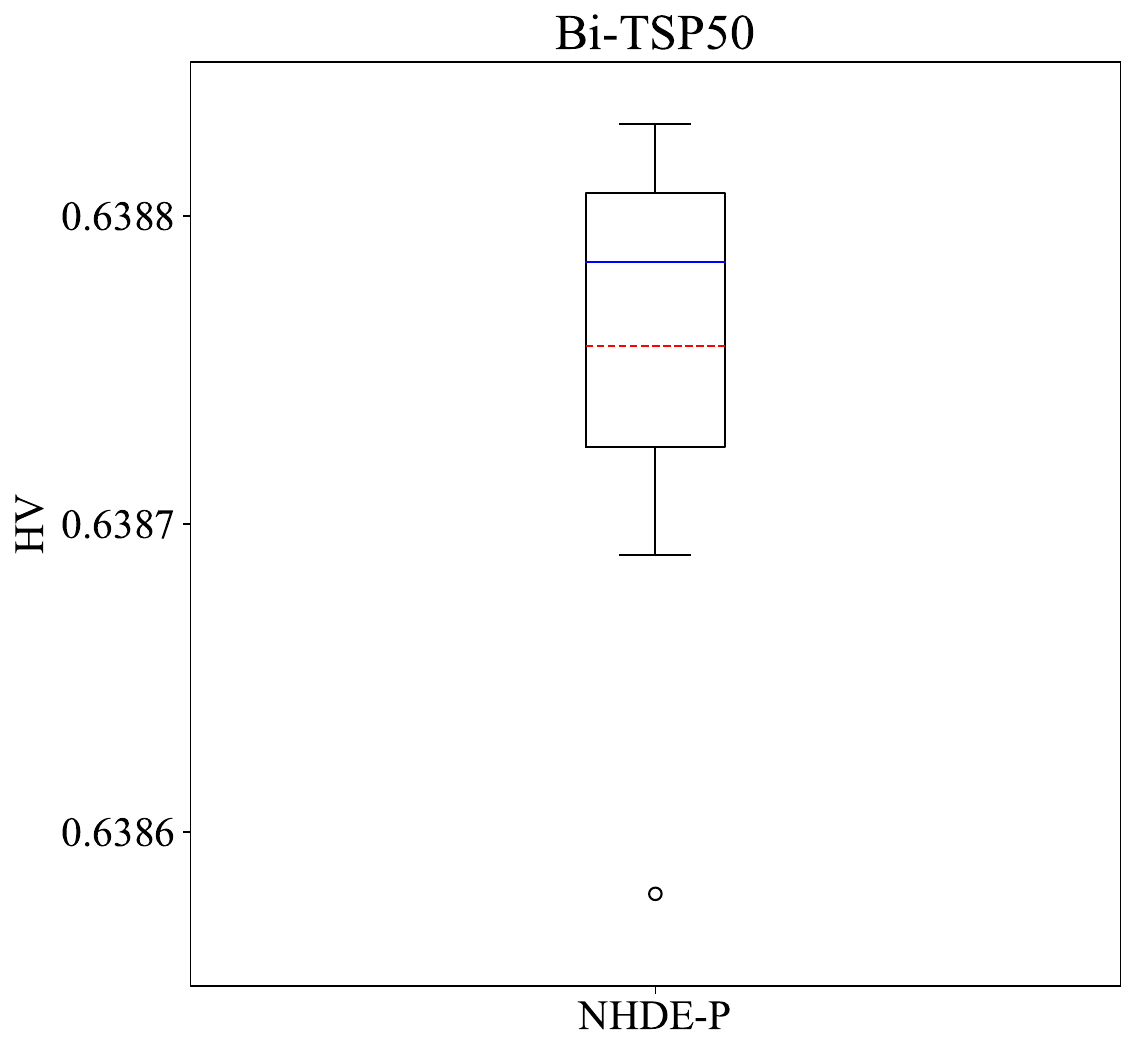}
		\label{fig:box}
	}
	\caption{Additional analysis. (a) Effect of diversity factors. (b) Effect of shuffled weights.}
	\label{fig:add}
\end{figure}

In the inference, diversity factors are linearly changed, which means different emphasis between the scalar objective and the HV indicator. We test other settings of the diversity factors, e.g., some fixed values. As shown in Figure \ref{fig:abla_w}, different settings of the diversity factors have almost no impact on the performance, except $\bm{w}^1=\dots=\bm{w}^N=(1,0)$ only emphasizing on HV. A possible reason is that the deep model is not good at learning the mapping from diversity factors to the complicated reward involving HV.

Recall that NHDE solves the subproblems dependently, and we simply use the shuffled weights. To study the effect of the random shuffle, we execute independent 10 runs. The boxplot of the results is presented in Figure \ref{fig:box}. As shown, the random shuffle of weights only exhibits a slight impact on the performance. A specialized order of the weights may raise the performance, but it is beyond the scope of this paper, which would be explored in the future.

We implicitly show the reduced duplicated solutions by the metric $|$NDS$|$, i.e., the number of non-dominated solutions. Empirically, more non-dominated solutions mean fewer duplicated solutions and fewer dominated solutions. Thus, the larger $|$NDS$|$ values of our method (especially in comparison to the other neural solvers) indicate that our method can produce a smaller number of duplicates to some extent. Furthermore, we add another evidence to verify that the indicator-enhanced DRL can hinder duplicated solutions. Specifically, we directly report the average number of duplicated solutions ($|$DS$|$) of our NHDE-P and NHDE-P without indicator (NHDE w/o I) on Bi-TSP50 in Table \ref{tab:ds}. As can be seen, our NHDE-P using the HV indicator can effectively guide the model to generate fewer duplicated solutions. Intuitively, our model is trained to construct new Pareto solutions different from existing ones in the Pareto front, which could achieve higher HV in the reward.

\begin{table}[!t]
  \centering
  \caption{Results for the number of duplicated solutions.}
    \begin{tabular}{lc}
    \toprule
    Method & $|$DS$|$ \\
    \midrule
    NHDE-P (40 wt.) & 196 \\
    NHDE w/o I (40 wt.) & 225 \\
    \bottomrule
    \end{tabular}%
  \label{tab:ds}%
\end{table}%

To further verify our results are statistically significant, we have conducted a Wilcoxon rank-sum test at a 1\% significance level for the results in all groups, which means our results are statistically significant. We additionally report the variances of the results in Table \ref{tab:var}. All methods have small variances of hypervolumes, where our NHDE-P achieves the stablest performance.

\begin{table}[!t]
  \centering
  \caption{Variances of the methods.}
    \resizebox{0.99\textwidth}{!}{
    \begin{tabular}{l|cc|cc|cc}
    \toprule
          & \multicolumn{2}{c|}{Bi-TSP20} & \multicolumn{2}{c|}{Bi-TSP50} & \multicolumn{2}{c}{Bi-TSP100} \\
    Method & HV    & Variance & HV    & Variance & HV    & Variance \\
    \midrule
    WS-LKH (40 wt.) & 0.6266  & $3.24\times 10^{-4}$ & 0.6402  & $1.59\times 10^{-4}$ & 0.7072  & $4.82\times 10^{-5}$ \\
    PPLS/D-C (200 iter.) & 0.6256  & $3.45\times 10^{-4}$ & 0.6282  & $1.72\times 10^{-4}$ & 0.6844  & $6.13\times 10^{-5}$ \\
    DRL-MOA (101 models) & 0.6257  & $3.31\times 10^{-4}$ & 0.6360  & $1.67\times 10^{-4}$ & 0.6970  & $5.09\times 10^{-5}$ \\
    \midrule
    PMOCO (40 wt.) & 0.6258  & $3.31\times 10^{-4}$ & 0.6331  & $1.64\times 10^{-4}$ & 0.6938  & $5.08\times 10^{-5}$ \\
    PMOCO (600 wt.) & 0.6267  & $3.28\times 10^{-4}$ & 0.6361  & $1.55\times 10^{-4}$ & 0.6978  & $4.71\times 10^{-5}$ \\
    NHDE-P (40 wt.) & 0.6286  & $3.19\times 10^{-4}$ & 0.6388  & $1.58\times 10^{-4}$ & 0.7005  & $4.76\times 10^{-5}$ \\
    \midrule
    PMOCO (40 wt. aug.) & 0.6266  & $3.26\times 10^{-4}$ & 0.6377  & $1.60\times 10^{-4}$ & 0.6993  & $4.99\times 10^{-5}$ \\
    PMOCO (100 wt. aug.) & 0.6270  & $3.28\times 10^{-4}$ & 0.6395  & $1.54\times 10^{-4}$ & 0.7016  & $4.81\times 10^{-5}$ \\
    NHDE-P (40 wt. aug.) & 0.6295  & $3.14\times 10^{-4}$ & 0.6429  & $1.52\times 10^{-4}$ & 0.7050  & $4.73\times 10^{-5}$ \\
    \bottomrule
    \end{tabular}%
    }
  \label{tab:var}%
\end{table}%



\end{document}